%% file: isca24-tender.tex
\documentclass[conference]{IEEEtran}
\IEEEoverridecommandlockouts
\usepackage{cite}
\usepackage{amsmath,amssymb,amsfonts}
\usepackage{algorithmic}
\usepackage{graphicx}
\usepackage{textcomp}
\usepackage{xcolor}
\def\BibTeX{{\rm B\kern-.05em{\sc i\kern-.025em b}\kern-.08em
    T\kern-.1667em\lower.7ex\hbox{E}\kern-.125emX}}

\input{macros}
\begin{document}

\title{Tender: Accelerating Large Language Models via Tensor Decomposition and Runtime Requantization\\
}

\author{\IEEEauthorblockN{Jungi Lee\IEEEauthorrefmark{1} \quad\quad Wonbeom Lee\IEEEauthorrefmark{1} \quad\quad Jaewoong Sim}
\IEEEauthorblockA{{Department of Electrical and Computer Engineering} \\
{Seoul National University}\\
\{jungi.lee, wonbeom, jaewoong\}@snu.ac.kr}
\thanks{\IEEEauthorrefmark{1}{Equal contribution}}
}

\maketitle

\thispagestyle{plain}
\pagestyle{plain}

\input{sections/abstract}


\input{sections/intro}
\input{sections/back}
\input{sections/algo}
\input{sections/arch}
\input{sections/eval}
\input{sections/discussion}

\input{sections/related}
\input{sections/conclusion}
\input{sections/ack}

\balance
\bibliographystyle{IEEEtranS}
\bibliography{refs}

\end{document}

%% file: macros.tex
\usepackage{fancyhdr}
\usepackage[normalem]{ulem}
\usepackage{xspace}
\usepackage{tikz}
\usepackage{subcaption}
\usepackage{dblfloatfix}
\usepackage{comment}
\usepackage{booktabs} 
\usepackage{graphics} 
\usepackage{pdfprivacy}
\usepackage{multirow}
\usepackage{xcolor}

\usepackage[font=small]{caption}

\usepackage{balance}

\usepackage[bookmarks=true,breaklinks=true,colorlinks,linkcolor=blue,citecolor=blue,urlcolor=black]{hyperref}

\usepackage{graphicx,pifont}
\let\oldding\ding
\renewcommand{\ding}[2][1]{\scalebox{#1}{\oldding{#2}}}

\newcommand{\name}{{Tender\xspace}}

\newcommand{\putsec}[2]{\section{#2}\label{sec:#1}}
\newcommand{\putssec}[2]{\subsection{#2}\label{ssec:#1}}

\newcommand{\secref}[1]{Section~\ref{sec:#1}}
\newcommand{\ssecref}[1]{Section~\ref{ssec:#1}}

\newcommand{\figref}[1]{Figure~\ref{fig:#1}}
\newcommand{\tabref}[1]{Table~\ref{tab:#1}}

\newcommand{\myparagraph}[1]{\vspace{0.02in}\noindent\textbf{#1}}

%% file: sections/abstract.tex
\begin{abstract}
Large language models (LLMs) demonstrate outstanding performance in various
tasks in machine learning and have thus become one of the most important
workloads in today's computing landscape. However, deploying LLM inference
poses challenges due to the high compute and memory requirements stemming
from the enormous model size and the difficulty of running it in the integer
pipelines.
In this paper, we present \name{}, an algorithm-hardware co-design solution
that enables efficient deployment of LLM inference at low precision. Based on
our analysis of outlier values in LLMs, we propose a decomposed quantization
technique in which the scale factors of decomposed matrices are powers of two
apart. The proposed scheme allows us to avoid explicit requantization (i.e.,
dequantization/quantization) when accumulating the partial sums from the
decomposed matrices, with a minimal extension to the commodity tensor compute
hardware.  Our evaluation shows that \name{} achieves higher accuracy and
inference performance compared to the state-of-the-art methods while also
being significantly less intrusive to the existing accelerators.
\end{abstract}

%% file: sections/intro.tex
\putsec{intro}{Introduction}

Large language models (LLMs) have demonstrated remarkable performance across a
variety of tasks in natural language processing, including machine translation,
sentiment analysis, and even generating human-like text, as evidenced by recent
applications such as OpenAI's ChatGPT and Google's
Gemini~\cite{wu:sch16,tho:de22,openai23,chatgpt,gemini}. The tremendous success
of LLMs can be largely attributed to their enormous model size, which has seen
substantial growth in recent years. For example, the first version of GPT,
which was introduced in 2018, had 117 million parameters, but the recently
released GPT-4 is rumored to contain more than a trillion parameters only after
five years.

LLM inference has now become one of the most important workloads in today's
computing landscape, but deploying and serving LLMs poses a unique challenge
because it requires a significant amount of compute and memory resources due to
the massive model size. Quantization~\cite{ami:seh21, det:lew22, zhe:rez22,
she:zhe20} is one of the most popular techniques to mitigate the resource
problem. By quantizing \emph{both} weights and activations in LLMs into low-bit
integers, we can accelerate compute-intensive operations such as matrix
multiplication while leveraging high-throughput integer tensor compute units in
modern GPUs or TPUs, in addition to benefiting from memory capacity and
bandwidth savings.

However, it is quite challenging to quantize the activations in LLMs, unlike
convolutional neural networks or small Transformer models. When the LLM scales
beyond a certain size (around 6.7B parameters), extremely large magnitude
values, compared to others, appear in a few feature dimensions of
activations~\cite{det:lew22}.  These \emph{outliers} increase the quantization
range, thereby necessitating the use of larger bit widths in LLMs compared to
other DNN models.

As such, there have been recent efforts to effectively quantize activations in
LLMs using low-bit integers in both software and algorithm-hardware co-design
works. However, most software-only works do not noticeably reduce the inference
time due to the overhead of complex algorithms or result in a significant
quantization loss at ultra low-bit precisions (e.g., 4
bits)~\cite{zhe:rez22,xia:lin22,lin:tan23,det:lew22}.
Also, prior works dealing with outliers via algorithm-hardware co-designs
require either mixed-precision/complex compute
units~\cite{par:kim18,det:lew22,zad:edo20,son:fu20} or custom/adaptive
datatypes that are not natively supported by commodity
hardware~\cite{guo:che22,guo:tan23,tam:yan20}.

In this paper, we propose \name{}, an algorithm-hardware co-design technique
that efficiently executes LLMs in the high-throughput integer pipeline without
the need of mixed-precision compute units or custom/adaptive datatypes.
The high-level idea behind \name{} is to split the \emph{activation} tensor
into several \emph{subtensors} along the feature/channel dimensions (e.g.,
columns in 2D), each of which contains the elements with a similar range of
values, effectively isolating the channels that contain outliers from the
others.  Each subtensor is then quantized with a different scale factor,
thereby reducing the quantization error for the entire activation tensor
compared to the conventional per-tensor (or per-row) quantization variants.

While this has the great potential to enhance model performance compared to
previous LLM quantization works, setting different scale factors for each
subtensor requires \emph{explicit} and \emph{costly} rescaling/requantization
(i.e., dequantization/quantization) when adding up the partial sums (i.e.,
outer products) from matrix multiplications of the subtensors.
Our key insight is that we can avoid the explicit requantization step by
setting the scale factors with \emph{power-of-two} relationships between the
decomposed subtensors and employing simple shifter logic in the tensor compute
units.
This approach provides us with two key benefits. First, it makes requantization
\emph{implicit}, being it performed along with matrix multiplication without
involving explicit floating-point operations, leading to negligible overhead
for rescaling when accumulating the partial sums.
Second, it enables higher model performance while being significantly less
intrusive to the conventional tensor compute units as it does not require
complex hardware to handle mixed-precision or custom datatypes, thereby
offering more flexible and practical applicability than other
algorithm-oriented schemes or outlier-aware architectures.
While the \name{} algorithm can be implemented in software, its full potential
is realized through a custom accelerator design that supports implicit
requantization, which we discuss in~\secref{arch}.

We apply our decomposed quantization technique to three representative LLMs
for evaluation. To measure the performance improvement, we implement a
cycle-level simulator that models the \name{} hardware with a detailed off-chip
memory timing model. We use simulation parameters based on our RTL
implementation, which is synthesized with a 28 nm process node.
Our evaluation shows that INT8 quantization via \name{} offers better model
performance than the state-of-the-art and retains comparable model performance
to the FP16 baseline (e.g., less than a 0.1 increase in perplexity for OPT-
6.7B, OPT-13B, and OPT-66B). In INT4 quantization, \name{} outperforms any
other outlier-aware post-training quantization (PTQ) techniques (up to
10988$\times$ lower perplexity). Also, the \name{} hardware achieves up
to an average of 2.63$\times$ speedup over the outlier-aware accelerators that
we evaluate.
In summary, this paper makes the following contributions:

\begin{itemize}
\item We propose \name{}, a PTQ approach for LLMs in pursuit of hardware
  performance as well as model accuracy. \name{} achieves high performance
  and accuracy without the need of mixed-precision compute units or custom
  datatypes even for low-bit quantization. 
\item We propose the ``power of 2'' channel decomposition rule, which
  effectively reduces the quantization error by harmonically working with LLM
  activation tensors.
\item We present the \name{} accelerator design, which overcomes performance
  challenges in splitting activations along the channels and enables
  implicit/runtime requantization with negligible rescaling overhead, with a
  minimal extension to the commodity tensor compute hardware.
\end{itemize}

%% file: sections/back.tex
\putsec{back}{Background and Motivation}

\putssec{}{Large Language Models}

\begin{figure}[t] 
\centering
\includegraphics[width=\linewidth]{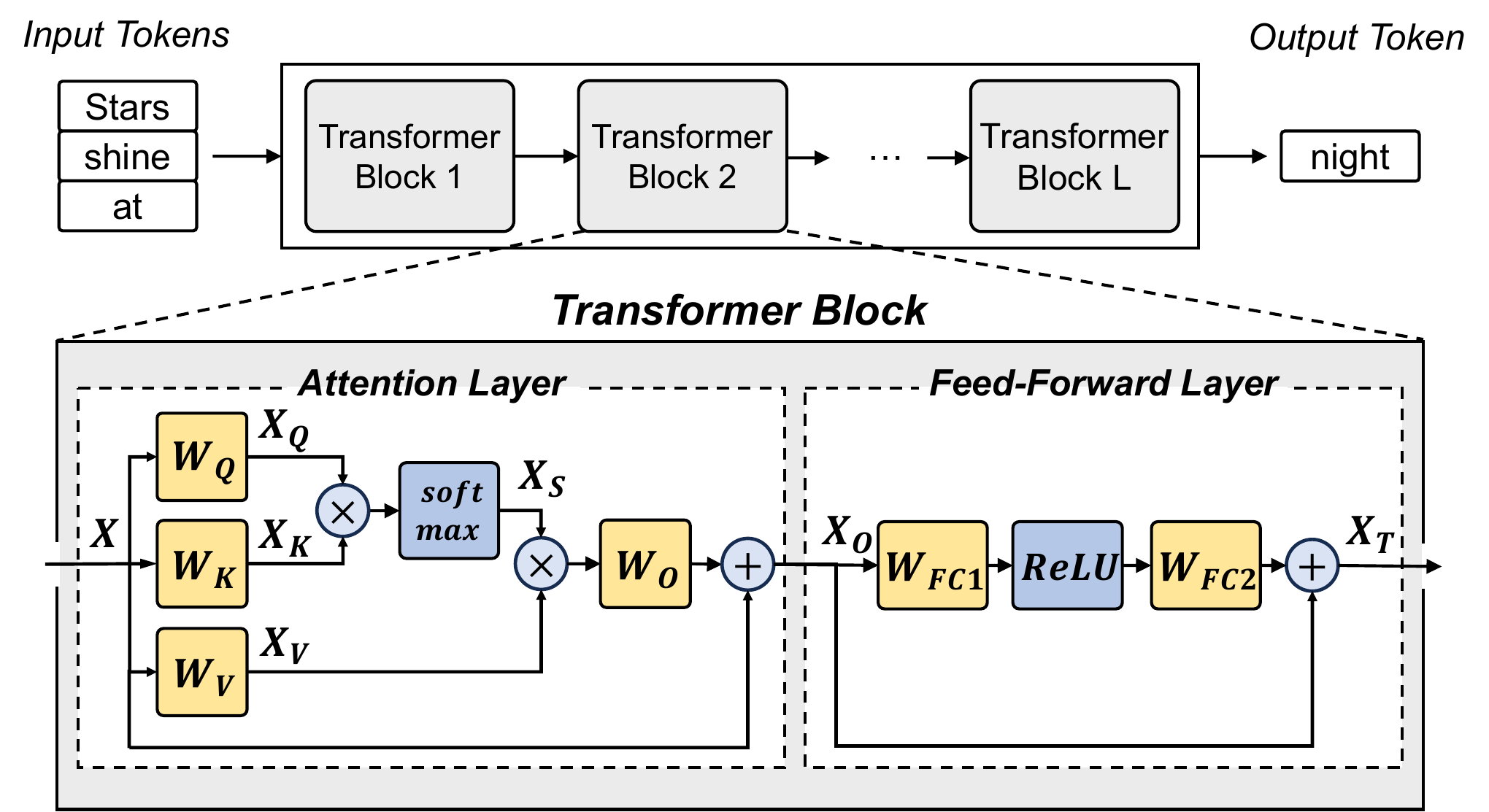}
\vspace{-0.15in} 
\caption{Illustration of the Transformer block architecture.}
\vspace{-0.15in}
\label{fig:transformer-block}
\end{figure}

In essence, LLMs build on a series of Transformer blocks~\cite{vas:sha17}, each
of which consists of the \emph{attention} and \emph{feed-forward} layers, as
illustrated in~\figref{transformer-block}.
In this section, for ease of understanding the subsequent sections, we briefly
describe the operation flow within the Transformer block using the terminology
and notation used throughout the paper.

In the attention layer, there are four weight matrices, which are used for
linear projections: query ($W_Q$), key ($W_K$), value ($W_V$), and output
($W_O$) projection matrices. Each projection matrix has a dimension of
$d_{model}$ ${\times}$ $d_{model}$, where $d_{model}$ is the dimension of an
input embedding vector (i.e., the number of features).
For an input sequence length of $n$ (i.e., $n$ tokens), the input ($X$) to the
attention layer is an $n$ $\times$ $d_{model}$ matrix.
Then, each of the query ($X_{Q}$), key ($X_{K}$), value ($X_{V}$) matrices of
the attention layer is computed by multiplying the input matrix ($X$) with
their respective weight matrices ($W$).
\begin{equation} 
  \small 
  \begin{aligned}
    X_{Q} = X \times W_{Q}; \quad X_{K} = X \times W_{K}; \quad X_{V} = X
    \times W_{V} \nonumber
  \end{aligned}
\end{equation}
After the QKV projection, we compute the attention score ($X_{S}$) and the
attention value ($X_{S}{\times}X_{V}$) using the projected matrices. 
Lastly, the output of the attention layer ($X_{O}$) is computed by multiplying
the attention value ($X_{S}{\times}X_{V}$) with the output weight matrix
($W_{O}$) along with a residual add ($X$).
\begin{equation} \small 
  \begin{aligned} X_{S} &=& \mathrm{softmax}(X_{Q} \times X_{K}^{T}) \\ X_{O}
  &=& X_{S} \times X_{V} \times W_{O} + X \nonumber \end{aligned}
\end{equation}

The feed-forward network (FFN) in the Transformer block takes as input the
output of the attention layer ($X_{O}$). It consists of two fully-connected
(FC) layers, thus there are two weight matrices, $W_{FC1}$ and $W_{FC2}$, each
of which has $d_{model}$ {$\times$} $h$ and $h$ {$\times$} $d_{model}$
dimensions. The output of the feed-forward layer ($X_{T}$) is computed using
the following equation.
\begin{equation} 
  \small 
  \begin{aligned} 
    X_{T} &=& \mathrm{ReLU}(X_{O} \times W_{FC1}) \times W_{FC2} + X_{O}
    \nonumber 
  \end{aligned} 
\end{equation}

In addition, there exists a layer normalization operation (LayerNorm) at the
start or the end of each attention and FFN layer, which we omit in the figure
for simplicity.
The Transformer block produces an output with the same dimensionality as the
input, which makes Transformer-based LLMs easily scalable by adjusting the
number of Transformer blocks.

\putssec{ol_in_llm}{Outliers in Large Language Models}

The state-of-the-art post-training quantization (PTQ)
methods~\cite{fra:ash23,lin:tan23} demonstrate that the weights in LLMs can be
effectively quantized to eight or even four bits via standard per-tensor (e.g.,
a scaling factor for each tensor) or grouping-based (e.g., a scaling factor for
$g$ consecutive weights in a tensor) quantization techniques without a
significant degradation in model performance.
In contrast, it is quite challenging to quantize \emph{activations} in LLMs
using the same methods due to the existence of \emph{outliers}, which refer to
the extremely large magnitude values compared to others within a tensor.

\begin{figure}[t] 
\centering
\includegraphics[width=\linewidth]{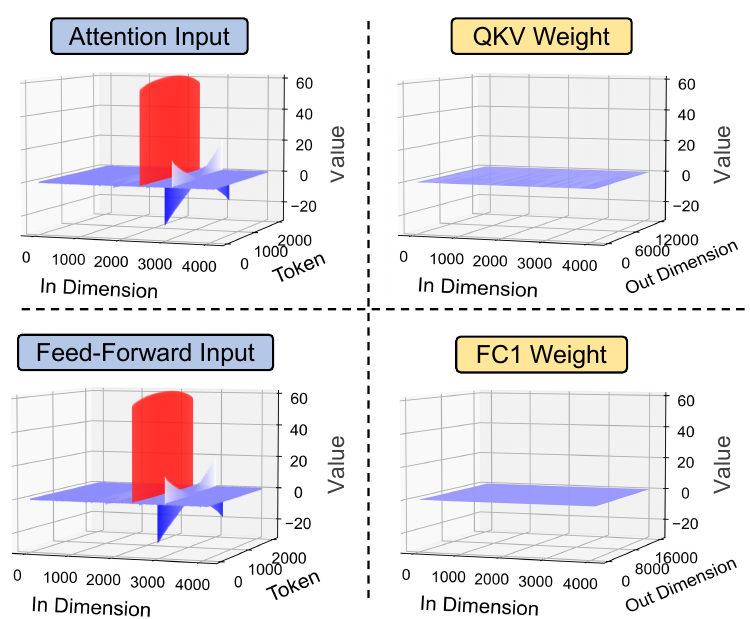}
\vspace{-0.20in} 
\caption{Values in the activation (left) and weight (right) tensors for the 
         attention and the first FC layers. The values are obtained from the 8th 
         layer in the OPT-6.7B model. The Q, K, V weight tensors are concatenated 
         along the out dimension for better visualization.}
\vspace{-0.20in} 
\label{fig:outliers} 
\end{figure}

\begin{figure}[t] 
\centering
\includegraphics[width=\linewidth]{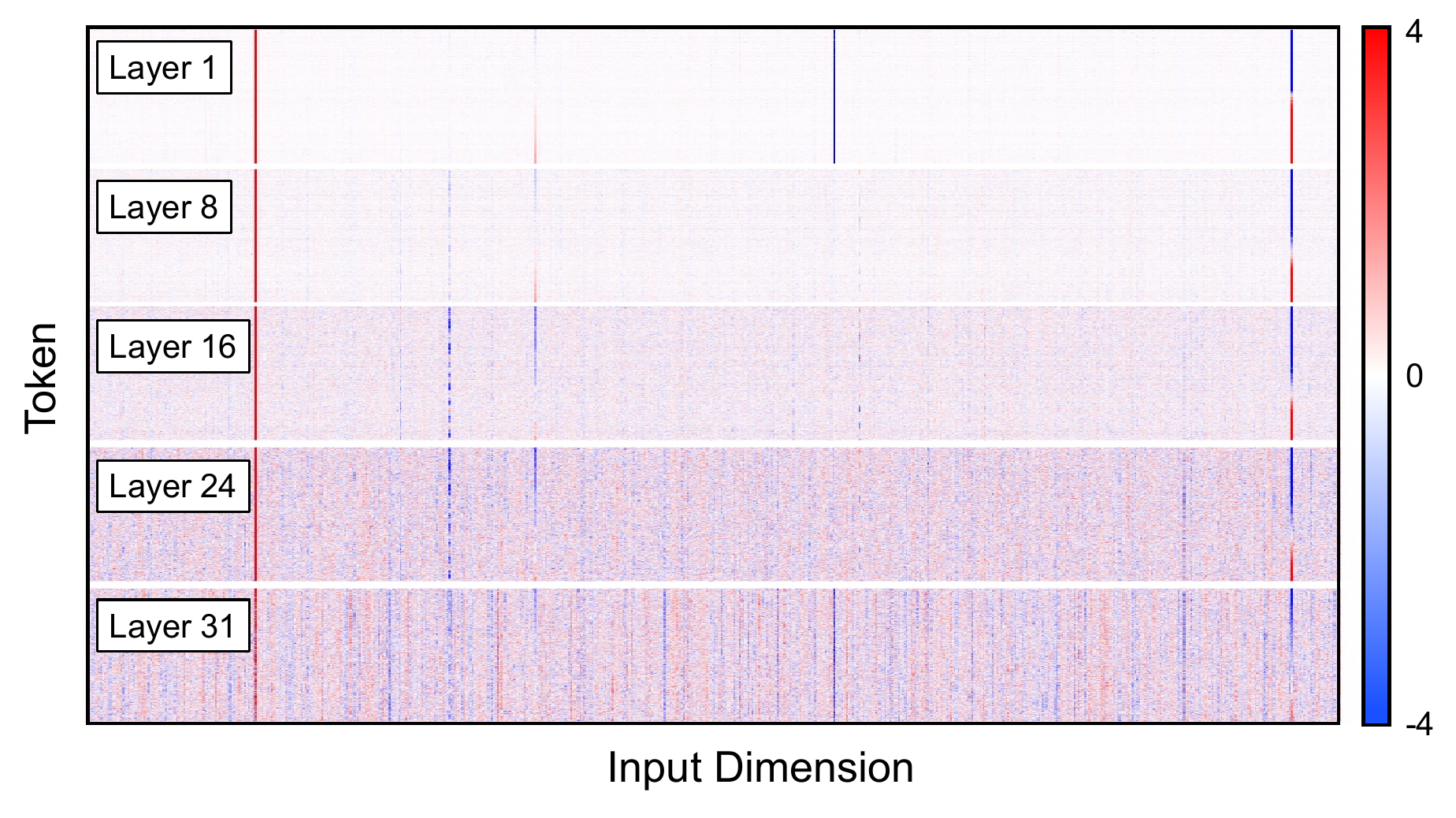}
\vspace{-0.25in} 
\caption{Heatmap of the attention input tensors for sampled layers in the 
         OPT-6.7B model. The values larger than 4.0 or smaller than -4.0 are 
         truncated to 4.0 or -4.0 for better visualization. We also only show 
         channels from 2300 to 3000 for a clearer view.}
\vspace{-0.20in}
\label{fig:cross_layer}
\end{figure}

Figure~\ref{fig:outliers} shows the values in several weight and activation
tensors. As depicted in the figure, the input activation tensors of the
attention and the first FC layers (i.e., $X$ and $X_{O}$) have significantly
large values in a few input (feature) dimensions, whereas the weight tensors
have a relatively similar range of values.  In principle, this makes it
difficult to quantize activations compared to the weight tensors in LLMs due to
the wide range of values, which we discuss further in the following sections.
Prior studies show that outliers are concentrated in the fixed channels of
activation tensors across the layers and batches~\cite{wei:yun22,det:lew22} due
to the model intrinsic, such as large LayerNorm weights in the fixed channels
across the layers.  
We also observe a similar trend, as shown in~\figref{cross_layer}.   
We see the presence of vertical red or blue lines for each attention
input tensor. This indicates that the outliers are typically at the channel
granularity (i.e., within a few specific channels).

\putssec{challenge_in_llmq}{Challenges in Quantizing Large Language Models}

\begin{table}[t]
  \caption{Model performance (perplexity) at different quantization granularities 
           for activation tensors. Lower is better.}
  \centering
   \resizebox{\linewidth}{!}{%
     \begin{tabular}{lcccc}
      \toprule
              Models & OPT-6.7B & OPT-13B & Llama-2-7B & Llama-2-13B \\
      \midrule
              FP16             & 10.86   & 10.13   & 5.47  & 4.88 \\
      \midrule
              INT8 per-tensor  & 26.73   & 4E+3    & 8.54  & 51.45 \\
              INT8 per-row     & 20.02   & 3E+3    & 5.58  & 4.94 \\
              \textbf{INT8 per-column}  &\textbf{10.87}   &\textbf{10.13}   &\textbf{5.48}  &\textbf{4.89} \\
      \midrule
              INT4 per-tensor  & 1E+6    & 9E+8    & 4E+4  & 2E+4 \\
              INT4 per-row     & 1E+6    & 1E+9    & 1E+3  & 5E+3 \\
              \textbf{INT4 per-column}  &\textbf{19.38}   &\textbf{14.60}   &\textbf{7.73}  &\textbf{6.47} \\
        \bottomrule
      \end{tabular}
    }
  \label{tab:act-quant}
\vspace{-0.20in}
\end{table}

\noindent\textbf{Preliminaries on Quantization.}
There have been various quantization techniques proposed over the past years.
However, the most commonly used one today is \emph{uniform integer}
quantization, as it is amenable to acceleration by the integer pipeline in
commodity hardware. As such, similar to prior works in LLM
quantization~\cite{xia:lin22,det:lew22}, this section focuses on uniform
symmetric quantization, which can be expressed as follows:
\begin{equation} 
  \small 
  \begin{aligned} 
    s = {{x_{max}}\over{2^{b-1}-1}}; \quad x_{q} = round({{x_{f}}\over{s}}),
    \nonumber
  \end{aligned} 
\end{equation}
where $s$ is a scale factor, $b$ is the bit width of a quantized value,
$x_{max}$ is the absolute maximum value, and $x_{f}$ and $x_{q}$ are a
floating-point value and the quantized one. Dequantization restores quantized
integer values to floating-point numbers by multiplying them with $s$.  To
mitigate the overhead of determining $x_{max}$ in activation during runtime,
most prior works employ \textit{static quantization}, which pre-computes the
scale factors using some calibration samples before runtime. 

Depending on how elements in a tensor are being quantized as a group, there can
be various levels of quantization granularity, including per-tensor, per-row,
and per-column quantization. In per-tensor quantization, all the elements
within the tensor share the same quantization parameter (i.e., a scale factor),
which simplifies the quantization process. Per-row or per-column quantization
shares the parameter at a row or column granularity to further reduce the
quantization error.

Table~\ref{tab:act-quant} shows the perplexity when we quantize the activation
at three different granularities. As shown in the table, the per-column
granularity (i.e., input/feature dimension) shows the best perplexity, as it
uses the quantization parameters for outliers that differ from others.
However, applying per-column quantization to activations poses challenges in
the modern GPU or TPU integer pipelines since each element needs scaling during
the reduction operations in matrix multiplication. Consequently, all prior LLM
quantization works employ per-row (per-token) or per-tensor quantization for
activations and per-column or per-tensor quantization for weights.

\myparagraph{Approaches to Handling Outliers in LLM Activations.}
Several algorithm-oriented PTQ works aim to handle outliers in LLM activations,
and two most closely related works are LLM.int8()~\cite{det:lew22} and
SmoothQuant~\cite{xia:lin22}.
%
LLM.int8()~\cite{det:lew22} employs a mixed-precision decomposition, where
outliers in activations are kept in FP16 precision, while the other values are
quantized to INT8. However, the mixed-precision decomposition leads to a
non-negligible performance overhead in performing matrix multiplication due to
the dequantization that involves floating-point operations~(\ssecref{quant}).
%
SmoothQuant~\cite{xia:lin22} addresses the quantization difficulty by partially
migrating it from the activations to weights. However, there exist
inefficiencies because it does not explicitly isolate outliers from normal
values, leading to a large quantization loss at ultra low-bit precisions
(\ssecref{acc}).

There are also several outlier-aware accelerators that use mixed precision to
quantize normal values into low bits while separately handling the outliers.
GOBO~\cite{zad:edo20} is a weight-only quantization scheme that quantizes
outlier weights using higher precision.  OLAccel~\cite{par:kim18} uses a few
16-bit MAC units alongside 4-bit MAC units to deal with outliers.
DRQ~\cite{son:fu20} employs a fine-grained detection algorithm to identify
sensitive regions in a tensor. All of these works use mixed precision,
requiring complex hardware and unaligned memory access.
OliVe~\cite{guo:tan23} is the most recent work, which quantizes outliers using
custom number representations. Although there is no mixed precision involved,
it prunes adjacent normal values and requires an encoder/decoder to support its
custom datatype.

\putssec{}{Challenges and Opportunities}
Intuitively, one can choose to split activation tensors along the reduction
axis (i.e., columns in 2D) and quantize each group with different scaling
factors instead of employing an impracticable per-column approach. We can then
represent the matrix multiplication via the channel grouping as follows:
\begin{equation}
\small
\begin{aligned}
    &P_{i} = \frac{X_{i} \times W_{i}}{s_{i}s_{w}}, &Y = \sum_{i=1}^{G} (s_{i}s_{w})\cdot P_{i},
\label{eqn:partial_sum_v1}
\end{aligned}
\end{equation}
where $s_i$ and $s_w$ are the scale factors of activation of group $i$ and
weight. $P_i$, $Y$, and $G$ denote the partial sum from group $i$, the final
resulting matrix, and the number of groups.

The above execution model still suffers from lower utilization of compute cores
due to smaller submatrices and frequent rescaling to accumulate partial product
$P_i$. Thus, to fully benefit from quantization while preserving the accuracy
of the model, we need to address the following challenge---splitting channels
along the reduction axis and grouping channels with similar ranges to isolate
outliers from the others while retaining the reduction axis to better utilize
compute cores.

Our intuition is that we can retain the reduction axis of matrix multiplication
by processing partial sums in a specific order and rescaling the accumulated
value before adding the next partial sum. Our execution model can be
expressed as the following equation:
\begin{equation}
\small
\begin{aligned}
    &A_1=P_1, \quad A_{i+1}=A_{i}\cdot\frac{s_{i}}{s_{i+1}}+P_{i+1},\\
    &\qquad \qquad Y = A_{G}\cdot(s_{w}s_{G}),
\label{eqn:partial_sum_v2}
\end{aligned}
\end{equation}
where $\mathrm{A_{i}}$ represents the results accumulated up to $i$th partial
sums. Note that Equations~\ref{eqn:partial_sum_v1} and \ref{eqn:partial_sum_v2}
are mathematically equivalent. 
For Equation~\ref{eqn:partial_sum_v2} to operate efficiently in hardware, the
rescaling needs to be performed by the integer MAC units while preserving
correctness.  We will use the term \textit{rescale factor} to denote
$s_{i}/s_{i+1}$ between the channel groups.
 
To this end, we propose \name{}, an algorithm hardware co-design technique
for quantizing large language models entirely into INT4/INT8 without using
mixed precision, custom datatypes, or re-training. The carefully designed
post-training quantization (PTQ) algorithm of \name{} decomposes channels to
isolate outlier channels from normal ones in activation tensors. At the same
time, \name{} enables rescaling inside the tensor compute units, so it
\emph{does not} require explicit requantization and thus fully utilizes the
integer pipelines. 
In the following sections, we discuss how the above rescaling can be done
inside the compute units with minimal extensions to hardware through the
software-hardware co-design of \name{}.

%% file: sections/algo.tex
\putsec{base-algo}{Algorithmic Implementation}

\putssec{overview}{Overview}

In this section, we introduce tensor decomposition and runtime requantization
of \name{}.  As discussed in~\ssecref{ol_in_llm}, outliers reside along the
channels in the activation tensors across the layers of LLMs.  Thus, we
decompose channels to minimize the quantization error by isolating the outlier
channels from the others.  We propose a ``Power of 2'' channel decomposition
rule that aligns well with the value distribution of activation tensors. Also,
we present runtime requantization that works harmonically with the
decomposition rule. It enables matrix multiplication between decomposed,
quantized activation tensors and linear symmetrically quantized weights without
involving floating-point operations but in a mathematically equivalent way. We
then present a walking example of the \name{} algorithm and optimizations at
the end of the section.

While the explanation in this section is based on INT8 quantized
activation-weight matrix multiplication, the same algorithm is applied to INT4
or activation-activation matrix multiplication (e.g., $X_{Q} \times X_{K}^{T}$
in a Transformer block).
Furthermore, \name{} can be easily extended to other bit widths (e.g., 5, 6,
7-bit integers) in the same way if the hardware supports such datatype
operations.  This is possible due to the standard and symmetric quantization of
\name{}, whereas other approaches typically need to define new custom
datatypes. 

\begin{figure*}[t] 
\centering
\includegraphics[width=\linewidth]{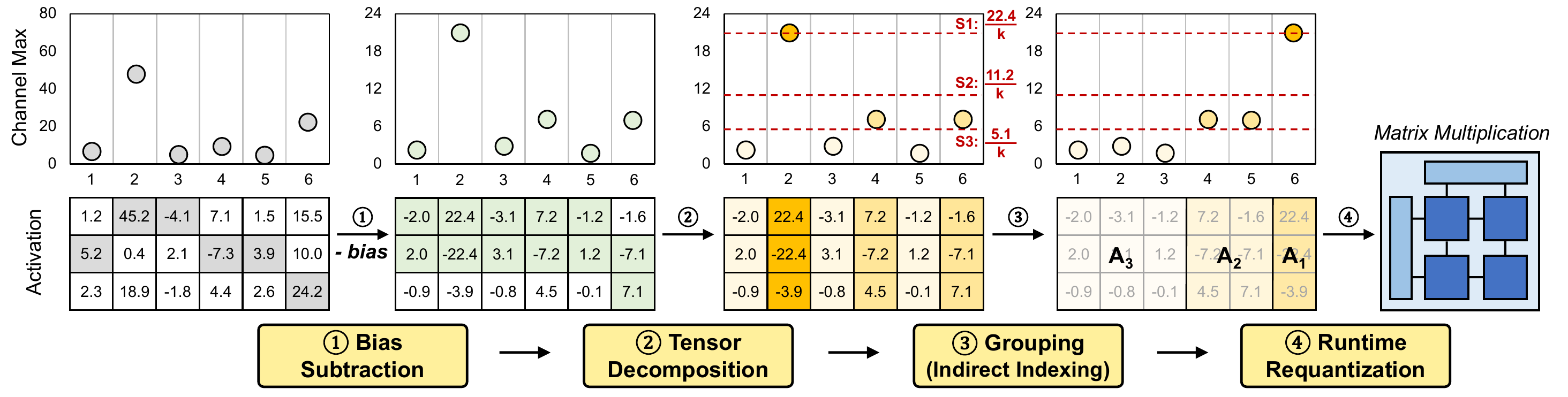}
  \caption{Decomposed quantization flow in \name{}.}
\vspace{-0.25in}
\label{fig:tender_flow} 
\end{figure*}

\putssec{quant}{\name{}: Decomposed Quantization}
\noindent\textbf{\name{} Computation Flow.}
As mentioned in~\ssecref{overview}, we decompose activation tensors throughout
the attention and feed-forward layers for quantization.
Figure~\ref{fig:tender_flow} shows our decomposed quantization strategy. First,
we compute the bias of each channel and subtract it from the activation tensor.
The bias is a similar concept to the zero-point used in asymmetric
quantization, and it is computed as the sum of the maximum and minimum values
divided by two. By subtracting the bias, \name{} ensures that the absolute
values of the maximum and minimum elements in the channel are equal, thus
optimizing the bit usage.

Then, to accommodate the presence of outliers in quantization, we decompose the
channels of the activation tensor into multiple groups of subtensors and use
separate scaling factors for each group. 
Through runtime requantization, we multiply the decomposed quantized activation
tensors and the linear symmetrically quantized weight without explicit
dequantization with negligible latency overhead. Finally, the bias multiplied
with weights is added to the output for mathematical correctness.
Note that all the weights can be quantized to INT8 before inference. Also,
channel decomposition, channel biases, and scale factors are pre-computed
during calibration.

\begin{figure}[t] 
\centering
\includegraphics[width=\linewidth]{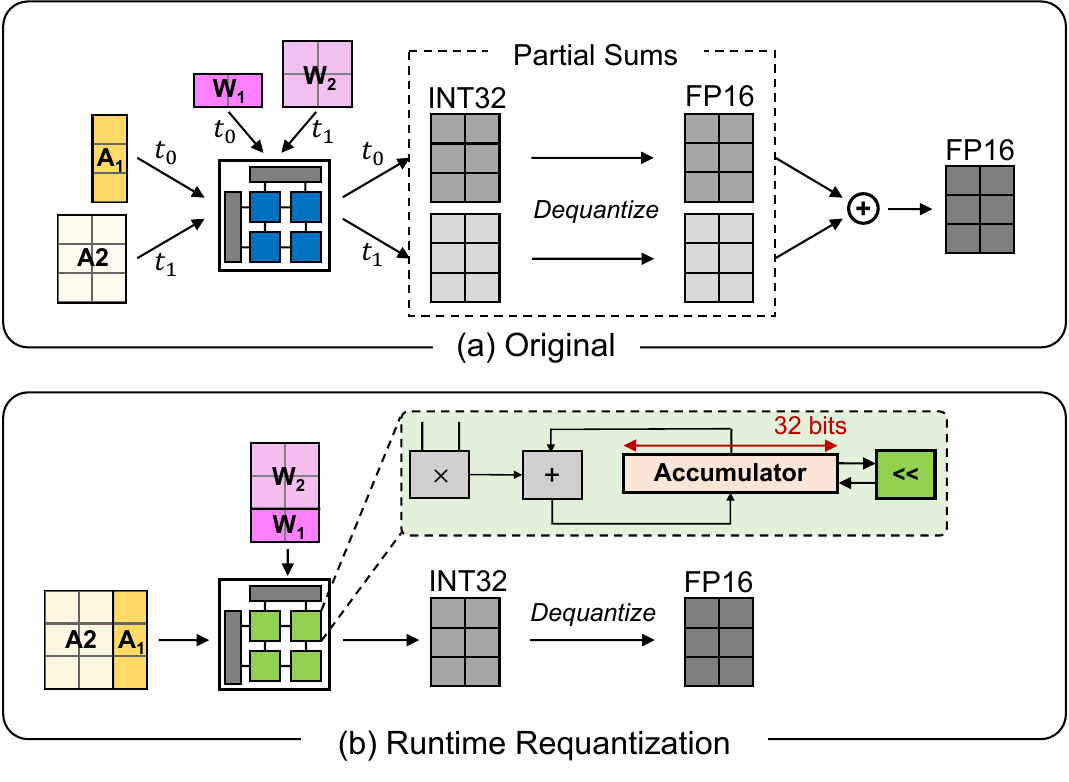}
\caption{Runtime Requantization. Compared to the original computation flow, 
         we retain the reduction axis length by shifting the accumulated 
         partial product sum between decomposed matrices.}
\vspace{-0.20in}
\label{fig:RR} 
\end{figure}

\myparagraph{Tensor Decomposition.}
\name{} decomposes channels by setting thresholds that are in powers of
integer relationships. Decomposition consists of three steps. First, \name{}
finds the absolute maximum values of each channel (CMax). \name{} also computes
the maximum value from CMax, which corresponds to the absolute maximum of a
given tensor (TMax). Then, we set the boundaries for splitting channels by
dividing TMax with the power of $\alpha$ and assign $\mathrm{i}$-th channel to
an appropriate group $\mathrm{g}$ satisfying the following equation:
\begin{equation}
    \small\mathrm{\frac{TMax}{\alpha^{g}} < CMax_{i} \le \frac{TMax}{\alpha^{g-1}}}, \quad g=1,2, ... G
\label{eqn:grp_idx}
\end{equation}

\noindent where $\mathrm{CMax_{i}}$ is the $\mathrm{i}$-th channel absolute
maximum.  This is simple classification, which is much faster than clustering.
Finally, every channel in group $g$ is quantized using the same scale factor
$\Large\mathrm{{\frac{TMax}{\alpha^{g-1}\cdot(2^{b-1}-1)}}}$.
If we choose an integer for $\alpha$ and compute channel groups with an
ascending group index, rescaling can be done using integer arithmetic by simply
multiplying $\alpha$ to an accumulated value. We use 2 for $\alpha$, so that
the requantization can be done with simple shifting.
More details about the reason why we use the power of $\alpha$ approach and 2
for $\alpha$ will be discussed in the following section.

\myparagraph{Power of 2.}
Here we explain our ``power of 2'' approach by answering the following three questions:
1) Why use classification? 
2) Why use coarse-grained thresholds for large values and fine-grained thresholds for small values?
3) Why use 2?

\textit{1) Why use classification?:} 
We can decompose channels in various ways. In the extreme scenario, we can use
individual scale factors for each channel, which can greatly reduce the
quantization error~\cite{xia:lin22,bon:nag21}. However, in this way, as
Figure~\ref{fig:RR}(a) shows, the number of operations including matrix
multiplication, dequantization, and addition between partial products increases
in proportion to the number of channels.  
In the case of OPT-6.7B~\cite{zha:rol22}, for example, it increases 4096 times
compared to the original computation without channel decomposition. As such,
grouping channels that have a similar range of values to share a scale factor
via clustering or classification is a more practical and scalable option.
Clustering computes how similar the value range is between every channel and
then groups the channels with a small distance. Classification is based on
pre-defined thresholds where the channels are binned to fit within the
threshold range.  Thus, clustering can group the channels more accurately than
classification but is not likely applicable at runtime (without complex
hardware) due to the large computational overhead. So, we choose to use
classification to make our algorithm simple and to be easily extended to
support runtime decomposition.

\textit{2) Why use coarse-grained thresholds for large values and fine-grained thresholds for small values?:}
The maximum quantization error is $0.5 \times \mathrm{(scale\ factor)}$ since
the maximum value of rounding error is $0.5$. The scale factor is set
proportional to the absolute maximum of the group. So, as the absolute maximum
of the group becomes larger, the quantization error grows linearly. Also,
considering that there exist multiple channels in a group, the quantization
error of a group grows linearly to the number of channels that are included in
the group.  We can express the quantization error of a group as follows:
\begin{equation}
    \label{quantization_error}
    Q_{\text{err}} \propto \text{Absolute Maximum} \times \text{Number of Channels} \nonumber
\end{equation}

Thus, for a group with a large absolute maximum, we need to minimize the number
of channels in the group. Intuitively, this can be viewed as isolating
quantization errors due to the large scale factor to only a few number of
channels. Similarly, for a group with a large number of channels, we need to
minimize the absolute maximum of the group. 
As Figure~\ref{fig:outliers} shows, only a few channels have large magnitude
values, and most of the channels have values near zero in the activation
tensor.
Thus, using coarse-grained thresholds (i.e., a large absolute maximum) for the
channels with large magnitude values does not hurt the overall accuracy because 
the group includes only a few channels. 
Meanwhile, using fine-grained thresholds (i.e., a small absolute maximum) for
the channels with small magnitude values can effectively minimize the accuracy
drop by using the small threshold near the value range of the channels.  Thus,
our approach can efficiently classify the channels with a minimum accuracy
drop.  
Note that, of course, it is possible to achieve better accuracy by using
fine-grained thresholds for both the channels with large magnitude values and
channels with small magnitude values.  However, dividing more channels incurs a
computation overhead, whereas using a fine-grained threshold for the channels
with large magnitude values has a minimal increase in accuracy due to the small
number of channels that are included in the group.

\textit{3) Why use 2?:}
Setting scale factors by dividing the maximum scale factor with the power of 2
has two key advantages. First, we can guarantee the lower bound of the
quantization level. When a channel is assigned to a quantization group, the
absolute maximum of the channel is at least larger than half of the threshold
of the group. Thus, even for the worst case, $n-1$ bits are utilized for
$n$-bit quantization.  Second, when the ratio between the scale factors is a
power of 2, we can efficiently compute the matrix multiplication involving
decomposed quantized activation tensors with a negligible latency overhead,
which we discuss in the following paragraph.
In summary, the benefit of the ``power of 2'' is that it not only enables
rescaling with integer arithmetic but also considers the alignment with the
channel distribution in activations.

\myparagraph{Runtime Requantization.}
Although tensor decomposition can effectively reduce the quantization error,
naively employing it requires an additional computation step to retain
functionality (Figure~\ref{fig:RR}(a)).  In a naive implementation, we must
decompose channels to generate multiple separate matrices and compute the
partial product of each group. Each partial product is dequantized using the
scale factor of each matrix and added up to the final result. This incurs an
additional overhead due to the increased number of floating-point operations.
Furthermore, decomposing the channels and computing the partial product of each
channel group lead to a shortened reduction axis. This results in
underutilization of compute cores especially in the systolic array
architecture.  To alleviate these inefficiencies and fully utilize the compute
cores, we propose \textit{Runtime Requantization}. 

Our intuition is that the systolic array accumulator has a sufficiently large
bit width, and we can safely \textit{requantize} the partial products with a
proper rescaling factor without clipping values due to limited bit width. 
In this way, we \textit{requantize} accumulated partial products without
involving explicit floating-point operations. We use a shifter residing next to
the accumulated sum for requantization since we use 2 as a rescaling factor.
Figure~\ref{fig:RR}(b) illustrates our runtime requantization approach.  We
perform matrix multiplication with a channel group that has the larger scale
factor first. Then, before computing matrix multiplication of the next group,
\name{} shifts the accumulated integer value by 1-bit.  After finishing the
matrix multiplication of all the groups, we dequantize the final result using
the smallest scale factor. This approach incurs a negligible latency overhead
for rescaling with a small hardware extension.
To reduce the overhead of determining a scale factor at runtime, we further
optimize our scheme to use calibration to pre-compute the scale factors and
biases of each channel offline. We also classify each channel into a group at
calibration time and only apply the metadata to perform quantized matrix
multiplications at runtime.

\myparagraph{A Walking Example.}
We explain how our decomposed quantization algorithm works with an example in
Figure~\ref{fig:tender_flow}, where there are six channels (channel IDs 1-6).
After subtracting the bias, each point represents the absolute maximum value of
each channel (CMax).
In the example, the 2nd channel has the largest CMax value among the channels.
Thus, we set the first (and largest) scale factor (\texttt{S1}) as the absolute
maximum value of the 2nd channel (i.e., 22.4) divided by $k=2^{b-1}-1$ (e.g., 127
and 7 for INT8 and INT4 quantization), where $b$ is the quantization bit width.  
We then set the subsequent quantization boundaries by dividing \texttt{S1} by
power of 2. For simplicity, we only consider three groups in the example, so
\texttt{S2} and \texttt{S3} would be 11.2 and 5.6 divided by $k=2^{b-1}-1$,
respectively.
Now, we classify each channel into one of the three groups. In the example, the
1st, 3rd, and 5th channels are assigned to the same group (\texttt{A3}),
sharing the same scale factor (\texttt{S3}). The 4th and 6th channels are also
classified into the same group (\texttt{A2}). {The 2nd channel is assigned
to another group (\texttt{A1}) with the largest scale factor (\texttt{S1}).}
After classification, six different channels are decomposed, so we can minimize
the quantization error while making the runtime requantization viable.  

\begin{figure*}[t]
  \centering
  \includegraphics[width=0.90\linewidth]{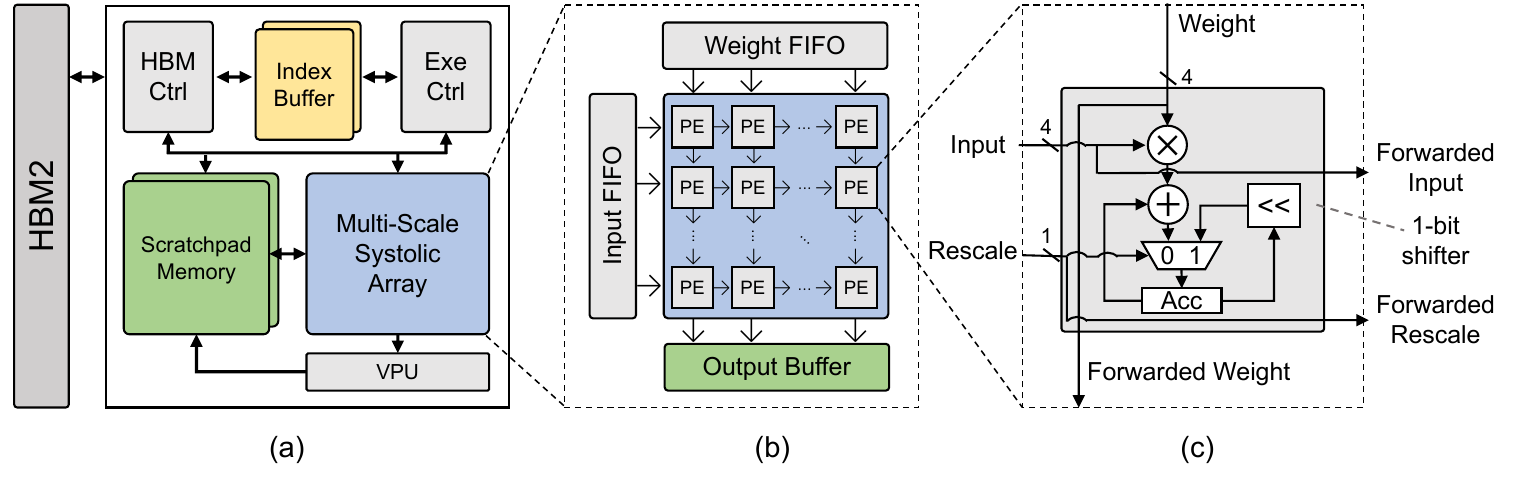}
  \vspace{-0.10in}
  \caption{(a) Overview of \name{} architecture.
           (b) Multi-Scale Systolic Array with FIFOs attached for skewing data.
           (c) Each PE is extended with a 1-bit shifter and a 1-bit control signal for rescaling.
            The PE updates an accumulated partial sum depending on the rescale signal.
            }
  \vspace{-0.15in}
  \label{fig:compute_overview}
\end{figure*}

\myparagraph{Optimization.}
We apply the \name{} algorithm to large language models and achieve similar or
better model performance compared to the existing quantization
works~\cite{guo:tan23, xia:lin22, guo:che22} on INT8 quantization.  We achieve
this while keeping the algorithm \emph{hardware-friendly} and the required
hardware \emph{minimal}. 
However, for INT4 quantization, the proposed algorithm inevitably leads to some
drops in model performance compared to FP16 due to its simplicity, most of
which can be easily recovered with conventional row-chunking techniques.

For example, we can divide the rows of the activation tensor into several
chunks and calibrate the bias and scale factor for each row chunk.
Figure~\ref{fig:outliers} shows that there exist not only inter-channel
variances but also \emph{intra-channel} variances. Thus, we may consider
intra-channel variances as well as inter-channel variances to group values more
effectively.
We observe that this can greatly improve the accuracy of our algorithm by
taking the characteristics of each row chunk into account. Because matrix
multiplication is typically performed in a tiled manner (i.e., tile-by-tile),
row chunking naturally fits the execution model with almost no additional
complexity. We use 256 as a row chunk size.  We also apply per-column weight
quantization and per-head activation quantization, which also incur negligible
calibration overhead. 

When optimizing with row chunking, channel grouping is applied to each row
chunk, and channel indices, biases, and scale factors are calibrated
\emph{independently} for each chunk offline. This makes the group size and
compute ordering differ \emph{between} row chunks.
From the hardware perspective, the row chunk size needs to be larger than the
systolic array dimension; otherwise, there can be underutilization of the
compute unit. This is because the systolic array computes rows and applies
runtime requantization at the granularity of the systolic array dimension.
While the on-chip buffer reuse decreases, it has a negligible impact on
performance with a reasonably large chunk size.
As mentioned, we choose 256 as a balance point, where accuracy remains close to
the baseline model due to fine-grained row grouping, and it is also
sufficiently larger than the systolic array dimension. 

%% file: sections/arch.tex
\putsec{arch}{Hardware Architecture}
We quantize the weights and activations in a Transformer block into INT4/INT8,
adopting the systolic array as the main computation module. The \name{}
architecture closely follows the conventional systolic design~\cite{jou:you17,
jou:kur23} with a simpler hardware configuration for brevity. In this section,
we address how our proposed channel decomposition and runtime requantization
are implemented in hardware with a minimal extension to the existing design. 

\putssec{}{Overview}
\figref{compute_overview}(a) shows an overview of the \name{} architecture.
Memories are HBM2, Scratchpad Memory, Output Buffer, and Index Buffer. The HBM
Controller manages data movement between the on-chip buffers and HBM2.  The
Execution Controller sends an address to the Scratchpad Memory to bring data
from the on-chip memory to the systolic array for computation. It also sends
control signals to the systolic array to manage its operations. The data is
computed in the Multi-Scale Systolic Array (MSA) and Vector Processing Unit
(VPU). Figures~\ref{fig:compute_overview}(b) and~\ref{fig:compute_overview}(c)
show our Multi-Scale Systolic Array architecture, which follows conventional
architectures with minimal extensions of a 1-bit rescale signal and a 1-bit
shifter inside the processing element (PE).

\begin{figure}[t] 
\centering
\includegraphics[width=\linewidth]{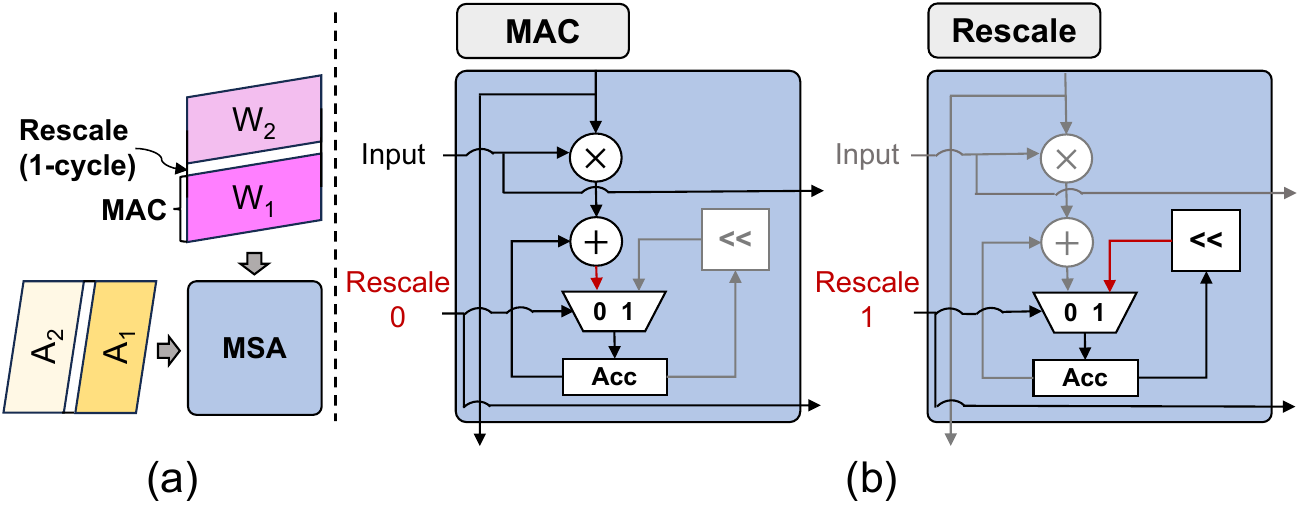}
\vspace{-0.25in} 
\caption{Execution model of the MSA. (a) A 1-cycle bubble is inserted between the decomposed matrices. 
        (b) Activated datapath inside PE during MAC and Rescale operation.
         }
\vspace{-0.20in}
\label{fig:ms-sys} 
\end{figure}

\putssec{ms-sys}{Multi-Scale Systolic Array (MSA)}
The main matrix multiplication operation occurs in our Multi-Scale Systolic
Array (MSA).  As illustrated in Figure~\ref{fig:compute_overview}(b), the MSA
is a 2-D mesh of PEs with FIFOs attached for skewing the inputs and weights. In
\name{}, we use a single 64$\times$64 systolic array with each PE executing a
4-bit MAC operation per cycle.  {When the model precision is INT8, 4 PEs are
grouped to perform 8-bit multiplication, with each PE handling either upper or
lower 4 bits of inputs and weights.} Note that we can scale the number of
systolic arrays similarly to the commercial design.

MSA is an output stationary compute module where a partial sum is accumulated
in each PE, with an extension of the 1-bit shifter attached to the accumulator
register. Figure~\ref{fig:ms-sys} shows how our MSA works on a series of
decomposed matrices with different scale factors.  During the computation of
each decomposed matrix, PEs perform normal MAC operations, and the accumulated
value is stored in the internal 32-bit register (Acc). When the PEs complete
matrix reduction for a decomposed matrix (e.g., $A_{1} \times W_{1}$), a
1-cycle bubble is inserted for the input and weight with a rescale signal set
(\figref{ms-sys}(a)). During the rescale operation, it requantizes the values
by shifting 1 bit of the accumulator register to the left.  Since each PE
finishes matrix reduction for a given decomposed matrix at different cycles,
the rescale signal is synchronized with the wavefront of the input and is
passed along to the next PE in the same row.  To generate rescale signals at
the right cycles, the Execution Controller has the metadata of the number of
channels executed, indices of tensor splitting points, and rescale factors (if
needed).

We choose our systolic array to be output stationary for two reasons. 
First, it allows us to easily extend the systolic array to handle arbitrary
rescale factors (i.e., other than $\alpha=2$) when needed.
Since the accumulator register is wired with the multiplier and shifter within
a PE, we can split the register into equal parts and multiply each
part with an arbitrary integer rescale factor that can come from the 4-bit
input or weight datapath.  
For example, with a given rescale factor $\alpha$ (e.g., 3) coming from the
input datapath, the accumulator is split into 8 parts (each with 4 bits).
Then, at each cycle, it is multiplied with the rescale factor, starting
from the part with the lowest bits, and its partial product is shifted to the
original position to be added to the resulting value.
The process is repeated for eight cycles to compute all the parts for
rescaling.

Second, the dataflow of the output stationary systolic array requires a minimal
hardware extension for rescaling. The output stationary systolic array can be
seen as an output value mapped to each PE, and a partial product is produced
and accumulated in the same PE every cycle. Thus, an additional 1-bit shifter
with control logic for each PE is enough for conventional hardware. For weight
stationary design, we need a shifter in the accumulator (which resides outside
of PE arrays) as well as in the PEs. Rescaling can be done as follows: 1)
Weights are loaded in the group order, and the PEs at the boundary of each
group are programmed to shift the partial product after MAC operations. 2) Each
corresponding accumulator shifts its value before adding an incoming partial
product. Note that although the above procedure requires slightly more changes
in hardware than output stationary, it is still a small extension to
existing hardware, and \name{} can also be implemented on the weight stationary
design.

\vspace{-0.02in}
\putssec{vpu}{Vector Processing Unit (VPU)}
The Vector Processing Unit (VPU) is a SIMD-style floating-point unit (FPU) that
operates on vector elements. It performs scaling of incoming INT32 results from
the Output Buffer (i.e., matrix multiplication results) into INT4/INT8 with an
optional activation (e.g., $\mathrm{ReLU}$, $\mathrm{GeLU}$) before storing it
back to the Scratchpad Memory.  It uses calibrated bias and scale factors,
which are computed before inference. VPU consists of 64 FPUs and internal
vector registers for pipelining. There are additional registers to buffer
scaling factors for quantization. Note that VPU also performs computation for
the softmax and LayerNorms in the Transformer block.

\putssec{}{Controllers \& Index Buffer}
The Execution Controller and HBM Controller operate independently during
computation to keep the MSA busy. The HBM Controller handles data transfers
between HBM2 and Scratchpad Memory, where the weights, inputs, and computed
outputs are stored at INT4/INT8 precision. The Execution Controller sends an
address to Scratchpad Memory and control signals (e.g., enable, rescale, and
done signal) to the MSA. 

\begin{figure}[t] 
\centering
\includegraphics[width=\linewidth]{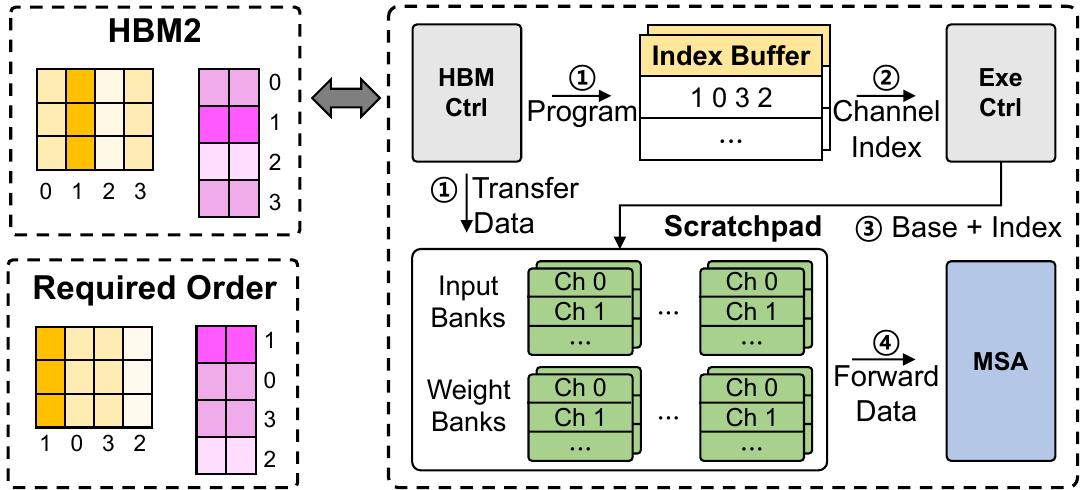}
\vspace{-0.15in} 
\caption{
        Dataflow of \name{}.
         }
\vspace{-0.15in}
\label{fig:idx-gen} 
\end{figure}

As discussed in~\ssecref{quant}, \name{} needs certain channels to be processed
before the others (e.g., channel order 1-0-3-2 in Figure~\ref{fig:idx-gen}). To
avoid explicit reordering of the data layout in memory, which incurs costly
read and write operations, we instead \textit{implicitly} reorder channels
through indirect indexing.  
Figure~\ref{fig:idx-gen} shows how channels are sent to the MSA in the required
order.  Specifically, we first store the computation order of the channel
indices in the Index Buffer, which is pre-determined at static time through
calibration (\raisebox{-0.2ex}{\ding[1.1]{172}} Program). The channel indices
are reused over the entire row group to amortize memory access overhead, and
the Index Buffer is double-buffered to hide memory access latency as it is on
the critical path.  While the HBM Controller also sends data from the off-chip
memory into the Scratchpad Memory (\raisebox{-0.2ex}{\ding[1.1]{172}} Transfer
Data), the Execution Controller looks up in the Index Buffer and obtains the
channel indices (\raisebox{-0.2ex}{\ding[1.1]{173}}) to generate an address for
the target channel to load (\raisebox{-0.2ex}{\ding[1.1]{174}}). Finally,
channels are sent to the MSA in the order of the required computation
(\raisebox{-0.2ex}{\ding[1.1]{175}}). 


\begin{table*}[t]
   \caption{INT8/INT4 PTQ results (perplexity) for large language models. 
            Lower is better. We omit LLaMA-65B due to the unacceptable increase 
            in perplexity in all other schemes except for \name{} in INT4 quantization.}
   \centering
   \resizebox{\linewidth}{!}{%
   \renewcommand{\arraystretch}{1.4}
      \begin{tabular}{cccccccccccccccccc}
        \toprule[1pt]
        \multirow{2}{*}{\textbf{Precision}} &\multirow{2}{*}{\textbf{Scheme}}  &\multicolumn{2}{c}{\textbf{OPT-6.7B}} &\multicolumn{2}{c}{\textbf{OPT-13B}} &\multicolumn{2}{c}{\textbf{OPT-66B}} &\multicolumn{2}{c}{\textbf{Llama-2-7B}} &\multicolumn{2}{c}{\textbf{Llama-2-13B}} &\multicolumn{2}{c}{\textbf{Llama-2-70B}} &\multicolumn{2}{c}{\textbf{LLaMA-7B}} &\multicolumn{2}{c}{\textbf{LLaMA-13B}}\\
                               &                   &Wiki          &PTB            &Wiki           &PTB            &Wiki          &PTB            &Wiki           &PTB            &Wiki          &PTB            &Wiki          &PTB            &Wiki      &PTB       &Wiki      &PTB   \\
        \bottomrule
            FP16               &Base               &10.86         &13.09          &10.13          &12.34          &9.34          &11.36          &5.47           &20.83          &4.88          &28.93          &3.32          &14.44          &5.68      &8.80      &5.09      &8.07 \\
        \bottomrule         
  \multirow{4}{*}{INT8}        &SmoothQuant        &\textbf{10.93}&13.21          &10.40          &12.53          &9.87          &11.71          &48.54          &1E+4           &447.52        &491.51         &17.30         &46.96          &27.85     &54.98     &16.02     &32.84  \\
                               &ANT                &19.72         &27.96          &4E+3           &3E+3           &3E+3          &3E+3           &8.79           &4E+4           &20.52         &152.01         &7.28          &36.18          &8.52      &13.41     &7.49      &10.85 \\
                               &OliVe              &\textbf{10.93}&13.23          &10.28          &12.41       &\textbf{9.43}    &11.41          &8.16           &30.12          &30.50         &26.16          &50.94         &245.09         &53.34     &113.48    &7.62      &10.76 \\
                               &\textbf{\name{}}&\textbf{10.93}&\textbf{13.14} &\textbf{10.17} &\textbf{12.39} &\textbf{9.43} &\textbf{11.40} &\textbf{5.77}  &\textbf{18.95} &\textbf{5.09} &\textbf{21.13} &\textbf{3.48} &\textbf{14.23} &\textbf{5.87}&\textbf{9.05}&\textbf{5.28}&\textbf{8.27}  \\
        \bottomrule       
  \multirow{4}{*}{INT4}        &SmoothQuant        &5E+4          &2E+4           &9E+3           &1E+4           &6E+4          &3E+4           &3E+5           &3E+5           &4E+4          &4E+4           &7E+4          &5E+4            &3E+5      &2E+5      &2E+5      &2E+5  \\
                               &ANT                &9E+3          &6E+3           &4E+4           &3E+4           &1E+4          &7E+3           &189.72         &2E+4           &165.19        &1E+3           &24.96         &155.92          &80.13     &109.21    &96.71     &247.65 \\
                               &OliVe              &50.83         &43.96          &35.76          &75.37          &6E+3          &4E+3           &44.24          &860.93         &1E+3       &\textbf{97.93}    &99.91         &216.53         &195.15    &359.43    &94.32     &181.69 \\
                               &\textbf{\name{}}&\textbf{13.56}&\textbf{16.28} &\textbf{16.43} &\textbf{19.92} &\textbf{12.38}&\textbf{14.01} &\textbf{36.47} &\textbf{114.44}&\textbf{55.08}   &208.76     &\textbf{13.43} &\textbf{50.66} &\textbf{23.85}&\textbf{38.09}&\textbf{13.68} &\textbf{28.24} \\
        \bottomrule[1pt]
      \end{tabular}
   }
   \label{tab:4/8b-acc}
   \vspace{-0.10in}
\end{table*}

\putssec{}{Scratchpad Memory \& Output Buffer}
As previously mentioned, all the inputs and weights are quantized into
INT4/INT8 and stored in the Scratchpad Memory. Without the need for mixed
precision storing, the memory access is aligned, and the addressing logic
becomes simpler. The Output Buffer stores the computation result from the MSA
in INT32 and sends them to the VPU for rescaling back to INT4/INT8, which is
also highly banked to match the compute throughput of the VPU.

%% file: sections/eval.tex
\putsec{eval}{Evaluation}

\putssec{method}{Experimental Methodology}

\noindent\textbf{Software Implementation.}
We implement our algorithm using PyTorch Hugging Face~\cite{wol:deb19}.
For evaluation, we use the Open Pre-trained Transformers (OPT)
suite~\cite{zha:rol22}, LLaMA~\cite{tou:lav23}, and Llama-2~\cite{tou:mar23}
with varying model sizes ranging from 6.7B to 70B to demonstrate the general
applicability of our proposed method.
We mainly evaluate language modeling tasks using WikiText-2~\cite{mer:xio16}
and Penn Treebank (PTB)~\cite{mar:kim94} datasets. We use perplexity as an
evaluation metric, which is a widely used one for autoregressive models; lower
perplexity means better model performance. 
To show that our algorithm works on the encoder-only model as well, we also
evaluate the accuracy of BERT-Large~\cite{dev:cha18} with the GLUE
benchmark~\cite{ale:ama19}.  
We use 128 samples from the Pile~\cite{gao:Bid20} validation set for
calibration to set scale factors, group indices, and a bias before runtime.

\myparagraph{Quantization Baselines.}
We compare the model performance of our quantization scheme with a variety of
outlier-aware PTQ works. For software-only quantization work, we compare with
SmoothQuant~\cite{xia:lin22}. SmoothQuant migrates the quantization difficulty
of activations to weights by scaling channels of inputs and
weights.\footnote{We use the original implementation of SmoothQuant. The
increase in model performance by enhanced SmoothQuant~\cite{neural-compressor}
was marginal while taking far longer calibration time in our experiments. The
model performance of enhanced SmoothQuant was still \emph{worse} than \name{}.}
We also compare our work with ANT~\cite{guo:che22} and OliVe~\cite{guo:tan23},
which target quantization under architectural support. OliVe employs
outlier-victim pair encoding, which sacrifices the normal value next to the
outlier to preserve the important outlier value. ANT proposes to adaptively use
different datatypes for different tensors. They both use custom data formats.

\myparagraph{Hardware Implementation.}
We implement \name{} in RTL with SystemVerilog and verify the functionality of
each component via RTL simulation. We report the area and power of \name{} by
synthesizing the components using a commercial 28 nm technology node with
Synopsys Design Compiler~\cite{synopsys}. On-chip SRAMs are also synthesized
from a commercial memory compiler with the same technology.
HBM2~\cite{JEDEC_HBM} is used as off-chip memory with the energy model from
FGDRAM~\cite{FGDRAM}. We also implement a cycle-level simulator with
Ramulator~\cite{kim:yan15} for DRAM timing to compare the performance of
\name{} and baseline accelerators.
The timing parameters of the simulator are set based on the RTL synthesis
results. \ssecref{perf} discusses the detailed configuration of \name{} with
the performance reported from our simulator. 

\myparagraph{Accelerator Baselines.}
We compare the performance and energy efficiency between \name{} and existing
quantization-based hardware accelerators:

\begin{itemize}

    \item OLAccel~\cite{par:kim18} proposes outlier-aware quantization, which
      {represents normal values in 4 bits and outliers in 8 or 16 bits. 
      In addition to the 4-bit normal PEs, OLAccel implements outlier PEs;
      outlier PEs perform mixed precision computation (e.g., 16-bit $\times$ 4-bit).}
    
    \item ANT~\cite{guo:che22} implements a systolic array with a decoder
      attached to the edge of the array to support various {formats 
      including custom datatypes. The decoder converts datatypes into the exponent and integer.}
      We implement an output stationary systolic array as it shows the best
      performance.
    
    \item OliVe~\cite{guo:tan23} also implements an output stationary systolic
      array with decoder logic {to decode datatypes including outlier-victim pairs
      into the exponent and integer}. 

\end{itemize}

For an iso-area comparison, we synthesize the MAC units and accumulators of
each accelerator and configure the number of PEs accordingly.  We extend the
baseline accelerators to use a 32-bit accumulator due to the large reduction
length of matrix multiplications in LLMs. Also, we set the same memory bandwidth
and on-chip buffer size for the accelerators, which are large enough to fully
utilize the compute core. We compare speedups in LLMs with a batch size of 1.
The input to output sequence length is set to 2048:1, following the speedup
evaluation in prior works~\cite{xia:lin22,guo:tan23}.
For the generation stage, \name{} still works and provides benefits by
decomposing the activation. However, the under-utilization issue of most
commercial accelerators (e.g., GPU, TPU) can be large as prior work points
out~\cite{hong:moon22}.  To mitigate this, there are ongoing studies on
batching decoding~\cite{yu:jeo22, she:zhe23}, and \name{} can work
synergistically with those schemes. 

\begin{table}[t]
   \caption {INT8/INT4 PTQ results (perplexity) across different sequence lengths. Lower is better.}
   \centering
   \resizebox{\linewidth}{!}{%
   \fontsize{10}{12}\selectfont
   \renewcommand{\arraystretch}{1.2}
      \begin{tabular}{cccccccc}
        \toprule[1pt]
        \multirow{2}{*}{\textbf{Precision}} &\multirow{2}{*}{\textbf{Scheme}} &\multicolumn{2}{c}{\textbf{2048}} &\multicolumn{2}{c}{\textbf{256}} &\multicolumn{2}{c}{\textbf{32}}  \\
        
                                   &                      &Wiki           &PTB            &Wiki           &PTB            &Wiki           &PTB \\
        \midrule
        FP16                       &Base                  &10.86          &13.09          &19.18          &22.00          &78.97          &103.42 \\
        \midrule          
        \multirow{5}{*}{INT8}      &SmoothQuant           &\textbf{10.93} &13.21          &\textbf{19.17} &22.14          &79.32          &\textbf{102.68} \\
                                   &ANT                   &19.72          &27.96          &48.43          &57.97          &396.01         &364.00 \\
                                   &OliVe                 &\textbf{10.93} &13.23          &19.24          &22.29          &79.69          &104.42 \\
                                   &\textbf{\name{} (all)}&10.98          &13.19          &19.31          &22.08          &78.93          &102.99 \\
                                   &\textbf{\name{}}      &\textbf{10.93} &\textbf{13.14} &19.28          &\textbf{22.06} &\textbf{78.81} &102.84 \\ 
        \midrule       
        \multirow{5}{*}{INT4}      &SmoothQuant           &5E+4           &2E+4           &5E+4           &2E+4           &4E+4           &2E+4 \\
                                   &ANT                   &9E+3           &6E+3           &8E+3           &6E+3           &6E+3           &3E+3 \\
                                   &OliVe                 &50.83          &43.96          &88.05          &113.53         &441.03         &371.73 \\
                                   &\textbf{\name{} (all)}&17.15          &23.25          &27.57          &30.58          &96.34          &118.85 \\        
                                   &\textbf{\name{}}      &\textbf{13.56} &\textbf{16.28} &\textbf{23.16} &\textbf{26.12} &\textbf{91.27} &\textbf{111.90} \\
                         
        \bottomrule[1pt]
      \end{tabular}
   }
   \label{tab:aa16}
\end{table}

\putssec{acc}{Language Model Performance}

\noindent\textbf{PTQ Performance on LLMs.}
We analyze the perplexity of PTQ for LLMs, which is the main target of
our work. \tabref{4/8b-acc} shows the perplexity under INT8 and INT4
quantization settings. The sequence length is set to 2048. For a fair
comparison with prior works, we disable the quantization of \name{} for matrix
multiplication between activations.
In INT8 quantization, \name{} consistently retains almost the same perplexity
from that of the FP16 baseline (less than a 6\% increase), while prior
works show up to a $1893\times$ increase in perplexity. Notably, \name{}
even outperforms the FP16 baseline in the Llama-2 models with the PTB
dataset. This can be due to the rounding in the quantization function. Since
the overall quantization error is quite low in INT8, rounding can prune out the
unnecessary small values, so that the model can instead focus on the important
ones.
In INT4 quantization, the outlier affects the model performance more
profoundly than in INT8 quantization. This is because quantizing outliers with
others leads to larger scale factors, and this effect becomes more pronounced
in INT4 PTQ which inherently has small quantization levels. Thus, isolating the
outlier channel is more important in INT4 quantization. \name{} shows far
better perplexity than others, which indicates that the channel
decomposition of \name{} can well separate the outlier channels from others and
classify the channels with similar ranges into the same group.

\myparagraph{Sequence Length Sensitivity.}
\tabref{aa16} shows the perplexity comparison between \name{} and prior
works for three different sequence lengths (2048, 256, 32) on
OPT-6.7B~\cite{zha:rol22}. Here we configure \name{} into two variants.
``\name{}'' disables the quantization for matrix multiplication between
activations for a fair comparison, while ``\name{} (all)'' quantizes all the
matrix multiplications in the Transformer block.
\name{} shows the best model performance for most of the quantization
scenarios. Notably, although \name{} (all) shows a slight increase in
perplexity, it even outperforms the prior works that do not quantize
matrix multiplication between activations in most cases. 
Furthermore, \name{} maintains the perplexity close to the FP16 baseline
even when the sequence length increases. This is due to the channel
decomposition which considers inter-channel variation and the row chunking
which handles intra-channel variation. As shown in the results, \name{} is more
robust than others while dealing with diverse scenarios of outlier values and
sequence lengths. Note that we use single calibration data attained from the
2048 sequence length for the evaluation across different sequence lengths.

\begin{table}[t]
  \caption {INT8/INT4 PTQ results (accuracy) on BERT-Large. Higher is better.}
  \centering
  \renewcommand{\arraystretch}{1.0}
  \linespread{1.3}
  \setlength{\aboverulesep}{0.1em}
  \setlength{\belowrulesep}{0.1em}
   \resizebox{\linewidth}{!}{%
   \fontsize{10}{12}\selectfont
      \begin{tabular}{cccccccc}
        \toprule
            \textbf{Precision}    &\textbf{Scheme}    &\textbf{CoLA}   &\textbf{SST-2}   &\textbf{MRPC}   &\textbf{STS-B}   &\textbf{QQP}      &\textbf{QNLI}  \\
        \midrule
            FP32                  &Base               &60.20           &93.12            &91.58           &89.94            &91.40             &92.33  \\
        \midrule          
            \multirow{3}{*}{INT8} &ANT                &59.16           &92.55            &77.99           &89.23            &89.66             &81.48 \\
                                  &OliVe              &\textbf{61.12}  &93.12            &91.33           &89.91            &91.42             &92.02 \\  
                                  &\textbf{\name{}}   &60.45           &\textbf{93.23}   &\textbf{91.55}  &\textbf{89.98}   &\textbf{91.43}    &\textbf{92.31}\\
        \midrule        
            \multirow{3}{*}{INT4} &ANT                &53.77           &90.60            &21.09           &85.93            &83.62             &60.86 \\
                                  &OliVe              &59.02           &92.09            &85.32           &87.43            &\textbf{89.72}    &\textbf{90.48}\\  
                                  &\textbf{\name{}}   &\textbf{61.78}  &\textbf{92.32}   &\textbf{89.42}  &\textbf{87.77}   &89.23             &90.29\\
        \bottomrule       
      \end{tabular}
      }
  \vspace{-0.10in}
  \label{tab:bert}
\end{table}

\myparagraph{Quantization Accuracy on BERT.}
\tabref{bert} shows the accuracy for INT8 and INT4 quantization on
BERT-Large~\cite{dev:cha18} with the GLUE benchmark~\cite{ale:ama19}. All
schemes in~\tabref{bert} quantize \emph{all} the matrix multiplications in a
Transformer block. Although the outliers of the BERT-Large are much smaller
than the ones of other large language models, \name{} outperforms other
baselines in many tasks. This indicates that our algorithm also benefits
encoder-only and relatively small models.

\begin{figure}[t]
  \centering
  \includegraphics[width=0.95\columnwidth]{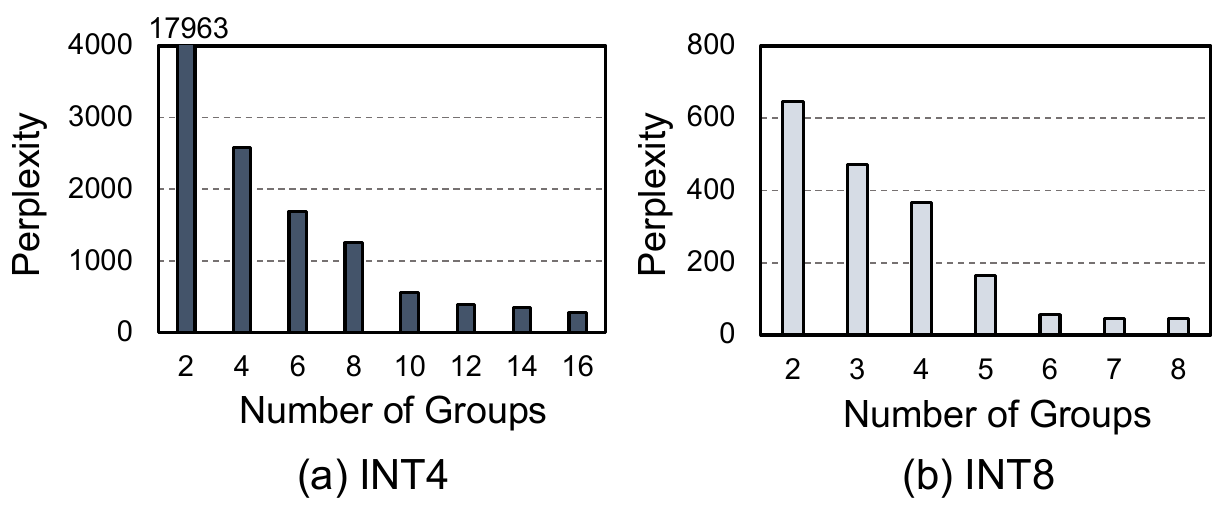}
  \caption{Perplexity for the different number of groups in (a) INT4 and (b) INT8 quantization. 
           Lower is better.}
  \vspace{-0.20in}
  \label{fig:n_group_sweep}
\end{figure}

\myparagraph{Multi-Scale Quantization.}
\figref{n_group_sweep} shows the perplexity on Llama-2-7B~\cite{tou:mar23}
while varying the number of groups for channel decomposition. We use the
PTB~\cite{mar:kim94} dataset and a fixed sequence length of 256. As we increase
the number of groups, the perplexity decreases rapidly for both INT4 and INT8
quantization. This shows that separating the channels only into two groups
(i.e., outlier channels and normal channels) is not enough, and decomposing the
channels into \emph{multiple} groups is necessary to achieve better model
performance. 
Note that naively adopting multi-scale quantization results in a frequent
interrupt during matrix multiplication. However, \name{} handles the
multi-scale quantization without interrupts by exploiting a minimally extended
systolic array.

\begin{table}[t]
  \caption{Area and power characteristics of \name{}.}
  \centering
    \resizebox{\columnwidth}{!}{%
      \begin{tabular}{llrr}
        \toprule
                    Component               &Setup                   &Area [mm$^2$]  &Power [W] \\
        \midrule
                    Systolic Array          &64$\times$64 PEs        &2.00           &1.09      \\
                    Vector Processing Unit  &64 FPUs                 &0.08           &0.02      \\
                    Input/Weight FIFOs      &64$\times$2             &0.05           &0.34      \\
        \midrule
                    Index Buffer            &2$\times$(16KB)         &0.23           &0.01      \\
                    Scratchpad Memory       &2$\times$(256KB)        &1.15           &0.13      \\
                    Output Buffer           &64KB                    &0.47           &0.01      \\
        \midrule
                    \textbf{Total}          &                        &\textbf{3.98}  &\textbf{1.60}   \\
        \bottomrule
      \end{tabular}
    }
  \label{tab:area_power}
\end{table}

\putssec{perf}{\name{} Performance}
\noindent \textbf{Area and Power.} 
\tabref{area_power} shows the architectural configurations of \name{}.
Functioning at the 1 GHz clock frequency, \name{} has an area of 3.98
$\mathrm{mm^{2}}$ with the peak power consumption of 1.60 W. The numbers
in the systolic array are MAC units and 32-bit accumulators combined. 
To match the compute throughput of the VPU, we also design the output buffer to
be highly banked while trading off area with throughput. 
We configure the PEs of the baseline accelerators to have the same area and
clock frequency as the ones in \name{} for performance and energy evaluation.

\begin{figure}[t]
  \centering
  \includegraphics[width=0.95\linewidth]{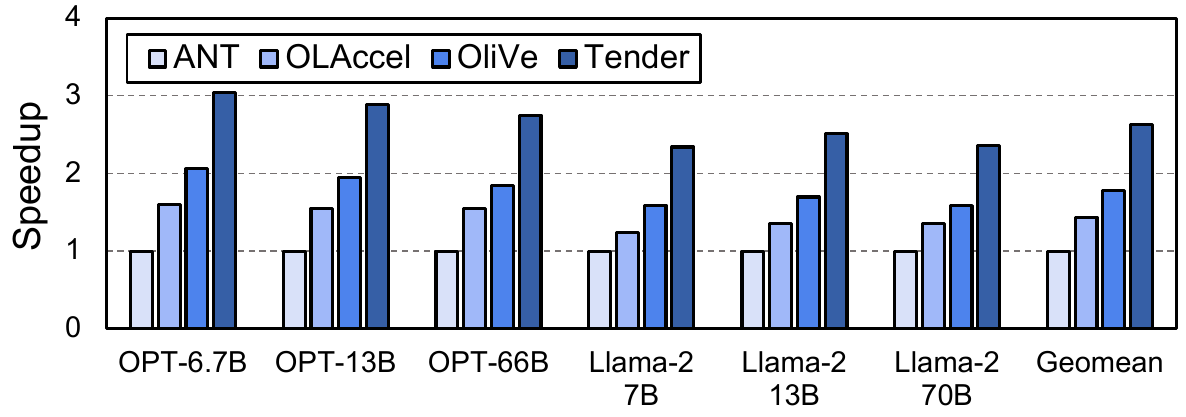}
  \vspace{-0.10in}
  \caption{Speedup comparison across the accelerators.}
  \vspace{-0.10in}
  \label{fig:speedup}
\end{figure}
\begin{figure}[t]
  \centering
  \includegraphics[width=0.95\linewidth]{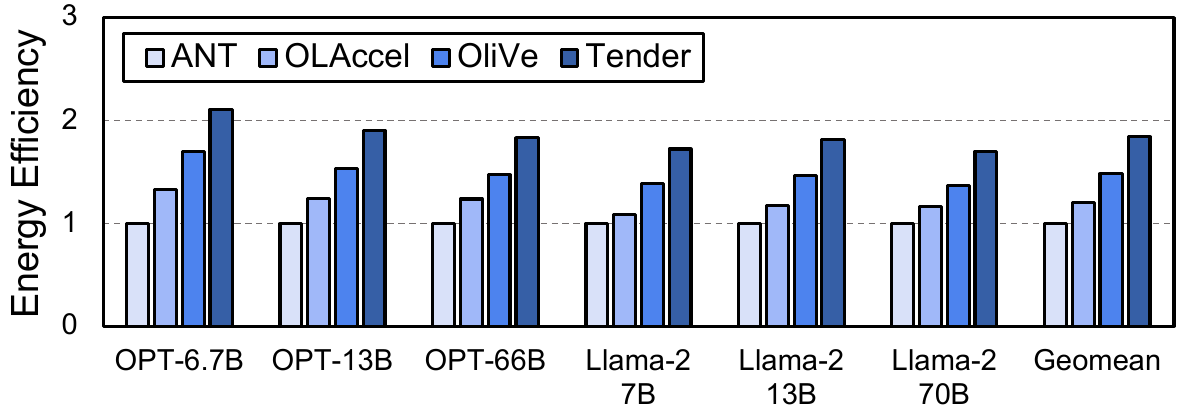}
  \vspace{-0.10in}
  \caption{Energy efficiency comparison across the accelerators.}
  \label{fig:energy}
\end{figure}

\myparagraph{Performance.} 
\figref{speedup} shows the speedup of \name{} and other accelerators over ANT
for the LLM models in~\ssecref{acc}; for brevity, we omit LLaMA as it shows the
results similar to Llama-2.
The source of speedups mainly comes from the careful algorithm-hardware
co-design of \name{}. Using single precision of INT4 with a hardware-friendly
tensor decomposition scheme, \name{} enables a simpler and denser
systolic array design with higher throughputs compared to others.  
In contrast, OLAccel has complex control logic and outlier PEs to support mixed
precision, and ANT and OliVe shift the multiplication result of the
integers with the exponent sum and require more hardware resources. 
\name{} only performs INT4 MAC operations and eliminates the need to handle
higher precision numbers.  
\name{} shows higher speedups compared to OliVe since OliVe computes using the
exponent and integer.  
ANT performs worse than other accelerators because most of the layers use 8-bit
precision to compensate for the quantization loss.  
Overall, \name{} achieves 2.63$\times$, 1.84$\times$, and 1.48$\times$ speedups
over ANT, OLAccel, and OliVe with better model performance and
minimal extensions to MAC units.

\myparagraph{Energy Efficiency.} 
\figref{energy} shows the energy efficiency of \name{} and the baseline
accelerators under the same off-chip memory and on-chip buffer size. 
The energy efficiency of \name{} mainly comes from a smaller memory size with
efficient hardware computation under the INT4 precision.
Compared with OLAccel, the energy efficiency comes from FIFO registers,
off-chip memory access, and compute units due to the shorter computation time.
For ANT, using mixed precision incurs more off-chip accesses and longer
computation latency, leading to higher energy consumption than \name{} and
OLAccel. Compared to OliVe, \name{} shows better efficiency due to the denser
systolic array design. 
Overall, \name{} shows 1.84$\times$, 1.53$\times$, and
1.24$\times$ higher energy efficiency than ANT, OLAccel, and
OliVe.  

%% file: sections/discussion.tex
\putsec{discussion}{Analysis and Discussion}

\begin{figure}[b]
  \centering
  \includegraphics[width=0.95\linewidth]
  {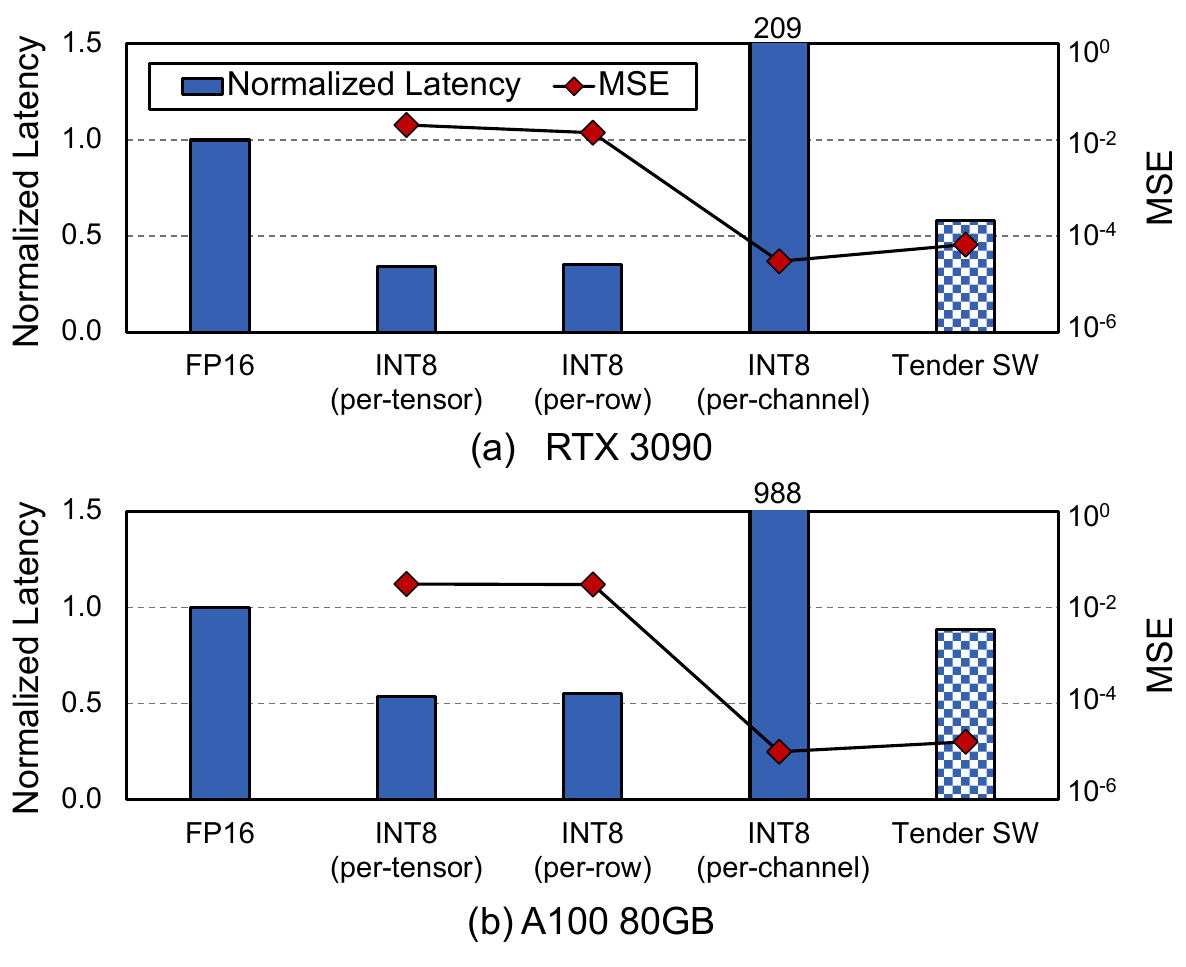}
  \caption{
            Comparison of Tender SW and other schemes on GPUs. 
            Latencies are measured in (a) RTX 3090 and (b) A100 80GB.
          }
  \label{fig:gpu_implementation}
\end{figure}

\putssec{}{GPU Implementation of Tender Decomposition}

\name{} uses standard INT4 and INT8 representations, thereby enabling
straightforward deployment to existing systems. To demonstrate the
non-intrusive characteristics of \name{}, we implement our quantization
algorithm on NVIDIA GPUs.
\figref{gpu_implementation} shows the normalized latency of \name{} software
and other quantization schemes in \ssecref{challenge_in_llmq}. All schemes are
implemented with CUTLASS INT8 GEMM kernel. We use OPT-6.7B for RTX 3090 and
OPT-66B for A100 to saturate compute units. 
For small models on A100, we observe that per-tensor INT8 GEMM exhibits similar
latency to FP16 due to compute underutilization and the relatively close tensor
core throughput between INT8 and FP16.
Latency and mean square error (MSE) are measured for each scheme with a sample
from the query projection in Layer 16.

As shown in the results, \name{} SW shows an MSE similar to the ``per-channel''
approach and provides slight performance benefits over FP16. However, it does
not realize its full potential (i.e., similar to ``per-tensor/-row'') due to
the need of explicit dequantization on GPUs. The overall GEMM execution also
takes longer due to the repetitive operations on smaller submatrices.
In addition, INT GEMM kernels for tensor cores require 128-bit aligned memory
access, which necessitates the padding of each subtensor (to the multiple of
16) before computation. Accelerators that support \name{} can avoid the
overheads and help achieve the full potential of \name{}. 

\putssec{}{Accelerators with Floating-Point Arithmetic} 

Variants of FP8 formats for DNN training/inference show good model
performance~\cite{mic:sto22}. Although using low-bit FP hardware could reduce
the impact of outliers, it is more area/energy \emph{inefficient} than integer
compute units~\cite{dar:zha23}.
MSFP~\cite{dar:lo20} uses a shared exponent to mitigate the inefficiency.
As shown in~\tabref{tender-bfp}, however, \name{} shows better performance than
MSFP. By default, MSFP12 uses an 8-bit shared exponent for 16 elements in a
\emph{row}, so the huge increase in perplexity likely comes from sharing
exponents between outliers and others. We modify the MSFP12 to use the shared
exponent for 8 elements in a column (MSFP12-OL). However, still \name{} shows
better model performance. This is likely because the intra-channel
variance of the outlier channel is more precisely represented in the integer
format than the MSFP. Thus, \name{} achieves better performance than
MSFP, while requiring much simpler hardware.

\begin{table}[t]
  \caption{PTQ perplexity of Tender and MSFP for WikiText-2.}
  \centering
    \resizebox{0.85\columnwidth}{!}{%
      \begin{tabular}{cccc}
        \toprule
                    Precision        &OPT-66B      &Llama-2-70B    &LLaMA-65B  \\ 
        \midrule
                    FP16             &9.34         &3.32           &3.56  \\
        \midrule
                    MSFP12           &7E+3         &74.61          &73.22 \\
                    MSFP12-OL        &56.69        &15.57          &26.11 \\
               \textbf{Tender-INT4}  &\textbf{13.38} &\textbf{13.43} &\textbf{9.30}  \\
        \bottomrule
      \end{tabular}
    }
  \label{tab:tender-bfp}
\end{table}

\putssec{}{Comparison with Microscaling (MX) Formats}

Shared Microexponents (SMX)~\cite{dar:zha23} and the subsequent Microscaling
(MX) format~\cite{mx-ocp} are recently proposed number representations that
employ multiple levels of scaling.\footnote{In this paper, we denote Shared
Microexponents~\cite{dar:zha23} as SMX to distinguish it from the Microscaling
(MX) formats~\cite{mx-ocp} endorsed by OCP.}
Similar to MSFP, they group elements in a block-based manner, but with
two-level scaling, where the scaling factors are constrained to powers of two.
SMX groups 16 elements with an 8-bit shared exponent, and two elements in a
block form a subgroup to share a 1-bit subscale factor. Similarly, MX groups 32
elements, but each element has also its \emph{own} exponent field in addition
to the 8-bit shared exponent.

\begin{table}[t]
  \caption{Accuracy for the lm-evaluation-harness zero-shot tasks used in~\cite{dar:zha23, dar:zha23-mx}. Higher is better.
                \name{} uses INT4.}
  \centering
  \resizebox{\columnwidth}{!}{%
    \begin{tabular}{c|cccc|cccc}
      \toprule
        \multirow{2}{*}{Tasks}    &\multicolumn{4}{c|}{OPT-6.7B}                                    &\multicolumn{4}{c}{LLaMA-7B}                       \\
                                  &FP32      &SMX4            &MXFP4           &\name{}             &{FP32}    &{SMX4}          &{MXFP4}         &{\name{}}      \\
      \midrule                                                                                                        
                  Hellaswag       &67.16     &26.94           &54.13           &\textbf{64.54}      &{76.20}   &{25.89}         &\textbf{67.51}  &{57.30}        \\
                  WIC             &48.12     &49.84           &\textbf{51.72}  &50.00               &{49.06}   &\textbf{50.00}  &{46.24}         &{49.53}        \\
                  Anli-r2         &34.40     &33.40           &33.90           &\textbf{34.20}      &{36.10}   &{33.40}         &\textbf{35.30}  &{35.20} \\         
                  Winogrande      &65.43     &50.12           &52.88           &\textbf{61.80}      &{70.01}   &{50.59}         &\textbf{62.35}  &{59.04}        \\     
      \midrule                                                                                                              
                  ARC easy        &60.02     &29.76           &44.57           &\textbf{56.82}      &{72.85}   &{27.78}         &\textbf{63.68}  &{58.50}        \\     
                  ARC challenge   &34.73     &23.46           &29.18           &\textbf{33.79}      &{44.71}   &{26.88}         &{35.49}         &\textbf{36.26} \\          
                  Lambada         &67.69     &00.02           &43.74           &\textbf{60.06}      &{73.61}   &{00.02}         &{56.65}         &\textbf{56.80} \\     
                  College CS      &34.00     &25.00           &25.00           &\textbf{34.00}      &{26.00}   &{23.00}         &{22.00}         &\textbf{28.00} \\     
                  Int. law        &37.19     &23.97           &\textbf{32.23}  &26.45               &{46.28}   &{29.75}         &{33.06}         &\textbf{33.88} \\     
                  Jurisprudence   &21.30     &\textbf{25.93}  &25.00           &21.30               &{36.11}   &\textbf{26.85}  &\textbf{26.85}  &{24.07}        \\      
      \bottomrule       
    \end{tabular}
    }
 \label{tab:lm-eval}
\end{table}

\tabref{lm-eval} compares the accuracy between employing \name{} and using SMX
and MX formats on OPT-6.7B and LLaMA-7B for the lm-evaluation-harness tasks
used in~\cite{dar:zha23,dar:zha23-mx}. 
For a fair comparison, we employ the same compute flow across the low-precision
formats while quantizing matrix multiplications into low precision and keeping
other element-wise operations as scalar floating-point formats as
in~\cite{dar:zha23-mx}.
The results show that Tender can provide better or comparable accuracy while
it builds on standard INT4 representations.
Note that SMX and MX need more customized compute units. For instance, each
element in an MXFP block is essentially a floating-point number, requiring
hardware that deals with FP computation.

The essence of \name{} involves setting scale factors that are powers of two
\emph{apart} between the groups (e.g., the \emph{ratios} of scale factors
between $1^\textrm{st}$/$2^\textrm{nd}$/$3^\textrm{rd}$ immediate neighbor
groups: $2^1, 2^2, 2^3$), which enables implicit rescaling with minimal
overhead (1-cycle). MX formats, however, merely use some power-of-two scale
factors, similar to MSFP. Thus, implicit rescaling by 1-bit shifting cannot
be achieved by simply using MX. As discussed in~\secref{base-algo}, the scale
factor of \name{} is also \emph{not} limited to powers of two; it can be any
real number. Additionally, MX formats follow a conventional approach of
grouping adjacent elements, whereas \name{} groups columns within similar
ranges while considering ease of computation. 

\putssec{}{\name{} on Output and Weight Stationary Dataflows}

While we present the benefit of \name{} based on the systolic array that
employs the output stationary dataflow, it is not a strict requirement as
discussed in~\ssecref{ms-sys}. 
For the generation stage in LLMs, we may consider batching inputs only up to
the number of rows of an output stationary systolic array due to compute and
energy efficiency, while we can batch more inputs than the systolic array
dimension for the weight stationary design.
If there are ample batching opportunities, weight stationary can be more
efficient since the output stationary design may incur idle time and additional
energy due to repeated weight loading. Conversely, when batching is limited,
such as by the memory size of large intermediate states (i.e., key-value
cache), output stationary could be as efficient as weight stationary since it
minimizes the movement of high-precision partial sums. 
Note that \name{} can be employed in either case and provides benefits.

\putssec{}{Channel Decomposition and Compute Utilization} 

Although channels are decomposed into multiple groups, \name{} can continuously
compute the groups in MSA without a major interrupt. This is because rescaling
can be done entirely in integer PEs (implicit) and only takes a single cycle
(i.e., runtime requantization). During runtime, skewed channel groups are
\emph{continuously} provided with a 1-cycle rescale signal, aligned with the
inputs between the groups (\figref{ms-sys}(a)), and each PE individually
rescales the accumulated value via 1-bit shifting; this feature is crucial as
the systolic array takes as input a skewed matrix. 
The original computation in~\figref{RR}(a) leads to large under-utilization of
compute units. 
\name{} does not suffer from this issue and \emph{preserves} the reduction axis
of the original input matrix, regardless of the number of groups or the group
size for the matrix. 
Thus, during offline calibration, \name{} only considers model
performance to determine the number of groups, where perplexity does not
decrease further as shown in~\figref{n_group_sweep}; it varies at some point
due to noise. 
Each channel is then classified into the corresponding group. This calibration
process naturally determines the size of each channel group.

\begin{figure}[t]
  \centering
  \includegraphics[width=0.95\linewidth]{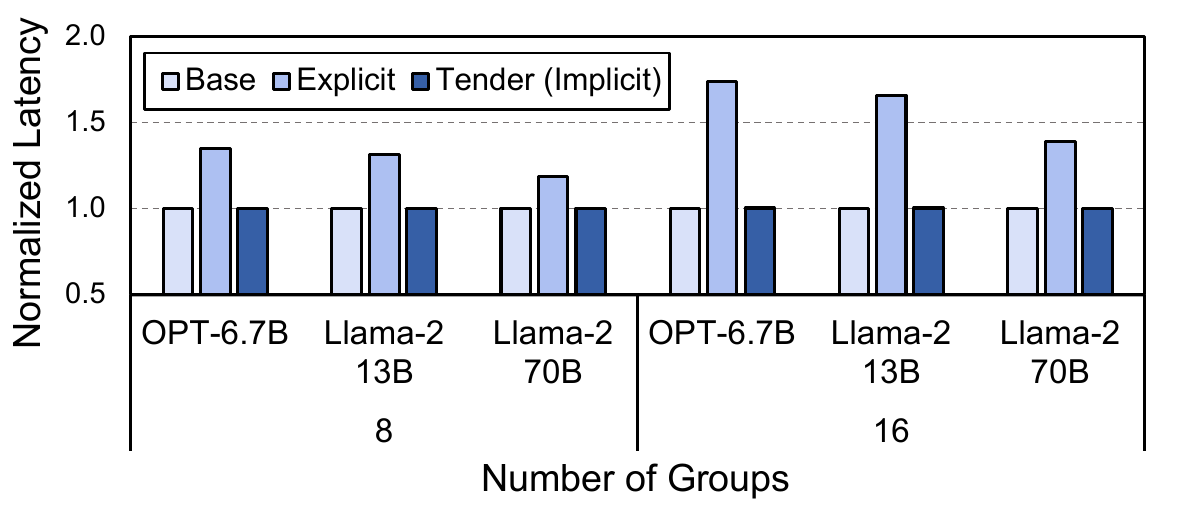}
  \caption{Comparison between implicit and explicit requantization.}
  \label{fig:explicit-overhead}
\end{figure}

\putssec{}{Impact of Implicit and Explicit Requantization}

Channel decomposition with \emph{explicit} requantization leads to lower
compute utilization due to the shortened reduction axis and also increases the
number of FP operations. To understand the benefit of implicit requantization
on \name{} hardware, \figref{explicit-overhead} presents the end-to-end
execution time when \name{} employs either implicit or explicit requantization,
which is normalized to per-tensor quantization (Base).
Note that larger models (e.g., Llama-2-70B) generally need more number of
groups to attain reasonable accuracy.
The results show that \name{} with explicit requantization greatly degrades
performance, by up to a {1.74}$\times$ slowdown over the baseline. Also, as the
larger number of groups (e.g., 16) further reduces the reduction axis, we
observe larger slowdowns compared to ones in the smaller number of groups.
On the other hand, \name{} (Implicit) offers almost the same execution time as
the baseline. This is because there is only a 1-cycle requantization overhead
for each group, so increasing the number of groups barely affects performance.
By employing implicit requantization, \name{} effectively minimizes the
overhead associated with channel decomposition.

%% file: sections/related.tex
\putsec{related}{Related Work}

\myparagraph{DNN Accelerators.}
Domain-specific accelerators for DNNs have been extensively studied over the
past decade~\cite{che:eme16,jou:kur23,rea:wha16,
eie,che:luo14,alw:che16,nur:ven17,fow:ovt18,eck:wan18,hong:moon22,qia:sai22}.
The processing units and dataflow of these accelerators are highly specialized
for DNN computation, leading to high performance and energy efficiency. Several
accelerators adopt near-memory processing to overcome the memory-bound
characteristics of specific types of DNN
workloads~\cite{ke:gup20,kwo:lee19,che:hai23,liu:lon23}. 
Other works also target sparsity in DNNs to skip ineffective
computation~\cite{par:rhu19,zha:du16,qin:ana20,alb:jor16,ham:lee21,lu:jin21,
wan:zha21,yaz:mor22}.
\name{} is orthogonal to these works and can be synergistically used with
conventional systolic array-based DNN accelerators.

\myparagraph{DNN Quantization.}
Quantization-aware training (QAT) trains the model under quantization to make
it adapt to quantization errors~\cite{shy:jav21,jac:kli18}. However, QAT is a
limited option due to the large model sizes, and thus post-training
quantization (PTQ) has been widely studied for
LLMs~\cite{xia:lin22,det:lew22,zhe:rez22}.
RPTQ~\cite{yua:niu23} employs K-means clustering to group activation channels
and applies asymmetric quantization at the granularity of a channel group.
However, each channel group needs to be computed one by one, leading to lower
compute utilization due to the smaller matrix sizes of each group. In addition,
all the partial products from each group need to be explicitly dequantized to
add up and obtain the final resulting matrix, which is costly. 
GPTQ (OPTQ)~\cite{fra:ash23}, AWQ~\cite{lin:tan23}, and QLoRA~\cite{det:pag23}
are the recent weight quantization works. GPTQ quantizes weights column by
column using the Hessian matrix to compensate the errors. AWQ scales weight
channels by observing outliers in activation tensors to reduce quantization
errors. QLoRA introduces the 4-bit NormalFloat datatype for block-wise weight
quantization by considering the distribution of values.

For quantization under architectural support, BitFusion~\cite{sha:par18}
proposes a bit-flexible architecture that can handle various precisions.
BiScaled-DNN~\cite{jai:ven19} introduces a new \textit{fixed-point} (FxP)
number format that employs two scale factors to represent values of small and
large magnitudes in a tensor. While it offers advantages over conventional FxP,
the heuristic for determining scale factors and the nature of its design would
not easily be applicable for a larger number of groups, unlike \name{}, which
is crucial for language model performance. Its \text{element-wise} metadata
also leads to more intrusive changes in conventional PEs.
ANT~\cite{guo:che22} adaptively uses a variety of datatypes, and
OliVe~\cite{guo:tan23} expresses outliers using specialized datatype to reduce
quantization errors. However, using custom or mixed datatypes makes it
challenging to be deployed for commodity hardware.
Compared to these works, \name{} offers a non-intrusive but effective solution
while building on the conventional number representation and datatype.

%% file: sections/conclusion.tex
\putsec{conclusion}{Conclusion}

Emerging large language models (LLMs) show remarkable model performance, but
quantization for scalability is challenging due to the presence of outliers in
the activation tensors. In this paper, we present \name{}, an
algorithm-hardware co-design that carefully considers both hardware performance
and quantization error. \name{} minimizes quantization errors by splitting the
activation tensors along the feature/channel dimension to separate the outlier
channels from the others. We address the runtime overhead of the channel
decomposition by introducing implicit requantization with the support of a
minimally extended systolic array. \name{} significantly improves PTQ
performance even for ultra low-bit quantization without mixed-precision
computation or custom datatypes. This work opens new possibilities for easier
and more efficient deployment of LLMs in a variety of practical scenarios.

%% file: sections/ack.tex
\section*{Acknowledgment}
We thank the anonymous reviewers for their valuable feedback.
This work was supported in part by a research grant from Samsung Advanced
Institute of Technology (SAIT) and by the artificial intelligence semiconductor
support program to nurture the best talents (No. RS-2023-00256081) supervised
by Institute for Information \& Communications Technology Planning \&
Evaluation (IITP).
The Institute of Engineering Research at Seoul National University provided
research facilities for this work.
Jaewoong Sim is the corresponding author.